\documentclass[sigconf]{acmart}
\AtBeginDocument{%
  }

\copyrightyear{2025}
\acmYear{2025}
\setcopyright{cc}
\setcctype{by}
\acmConference[KDD '25]{Proceedings of the 31st ACM SIGKDD Conference on Knowledge Discovery and Data Mining V.2}{August 3--7, 2025}{Toronto, ON, Canada}
\acmBooktitle{Proceedings of the 31st ACM SIGKDD Conference on Knowledge Discovery and Data Mining V.2 (KDD '25), August 3--7, 2025, Toronto, ON, Canada}
\acmDOI{10.1145/3711896.3736932}
\acmISBN{979-8-4007-1454-2/2025/08}



\graphicspath{{./figures/}}

\usepackage{xspace}
\usepackage{graphicx}
\usepackage{amsmath}
\usepackage{geometry}
\usepackage{subcaption}
\usepackage[ruled,linesnumbered]{algorithm2e} 
\usepackage{algorithmic}
\usepackage{fancyvrb}
\usepackage{balance}

\usepackage[inline]{enumitem}
\usepackage[show]{chato-notes}

\usepackage{balance}

\usepackage{xcolor}

\newcommand{\prompt}{\ensuremath{q}\xspace}
\newcommand{\generated}{\ensuremath{g}\xspace}
\newcommand{\content}{\ensuremath{t}\xspace}
\newcommand{\opinion}{\ensuremath{s}\xspace}

\newcommand{\llm}{\ensuremath{\textrm{LLM}_\theta}\xspace}
\newcommand{\llmm}[1]{\ensuremath{\textrm{LLM}_{#1}}\xspace}

\newcommand{\readability}{\ensuremath{f}\xspace}
\newcommand{\reward}{\ensuremath{r}\xspace}
\newcommand{\finalreward}{\ensuremath{\mathcal{R}}\xspace}
\newcommand{\network}{\ensuremath{\mathcal{G}}\xspace}

\newcommand{\maxiterations}{\ensuremath{k}\xspace}
\newcommand{\klthreshold}{\ensuremath{\tau}\xspace}

\newcommand{\utilitymodel}{\ensuremath{\mathcal{S}}\xspace}
\newcommand{\propagationprotocol}{\ensuremath{\mathcal{M}}\xspace}

\newcommand{\spara}[1]{\smallskip\noindent{\bf #1}}

\newcommand{\pos}{\ensuremath{u_{\mathrm{L}}}}

\newcommand{\added}[1]{\textcolor{black}{#1}}

\begin{document}

\title[Engagement-Driven Content Generation with Large Language Models]{Engagement-Driven Content Generation\\ with Large Language Models}

\author{Erica Coppolillo}
\affiliation{%
\institution{University of Calabria}
\department{Department of Computer Science}
\country{}
}
\affiliation{%
\institution{ICAR-CNR, Rende, Italy}
\country{}
}
\email{erica.coppolillo@unical.it}

\author{Federico Cinus}
\affiliation{%
  \institution{CENTAI}
  \city{Turin}
  \country{Italy}
}
\email{federico.cinus@centai.eu}

\author{Marco Minici}
\affiliation{%
  \institution{ICAR-CNR}
  \city{Rende}
  \country{Italy}
}
\email{marco.minici@icar.cnr.it}

\author{Francesco Bonchi}
\affiliation{%
  \institution{CENTAI, Turin, Italy}
 \country{}
}
\affiliation{%
  \institution{Eurecat, Barcelona, Spain}
 \country{}
}

\email{francesco.bonchi@centai.eu}

\author{Giuseppe Manco}
\affiliation{%
  \institution{ICAR-CNR}
  \city{Rende}
  \country{Italy}
}
\email{giuseppe.manco@icar.cnr.it}


\begin{abstract}
Large Language Models (LLMs) demonstrate significant persuasive capabilities in one-on-one interactions, but their influence within social networks, where interconnected users and complex opinion dynamics pose unique challenges, remains underexplored. This paper addresses the research question: \emph{Can LLMs generate meaningful content that maximizes user engagement on social networks?}

To answer this, we propose a pipeline using reinforcement learning with simulated feedback, where the network's response to LLM-generated content (i.e., the reward) is simulated through a formal engagement model. This approach bypasses the temporal cost and complexity of live experiments, enabling an efficient feedback loop between the LLM and the network under study. It also allows to control over endogenous factors such as the LLM's position within the social network and the distribution of opinions on a given topic. Our approach is adaptive to the opinion distribution of the underlying network and agnostic to the specifics of the engagement model, which is embedded as a plug-and-play component. Such flexibility makes it suitable for more complex engagement tasks and interventions in computational social science.

Using our framework, we analyze the performance of LLMs in generating social engagement under different conditions, showcasing their full potential in this task. The experimental code is publicly available.\footnote{\url{https://github.com/mminici/Engagement-Driven-Content-Generation}}

\end{abstract}


\maketitle \sloppy

\section{Introduction}\label{sec:intro}

Large Language Models (LLMs) have recently gained a great deal of attention, especially for their capabilities in one-to-one interactions. Researchers have deeply investigated  their skills and limits, comparing LLMs to humans in one-on-one interactions such as, e.g., their abilities in gaming settings~\cite{jacobconsensus,fontana2024nicer}.
A recent research direction empirically investigates the susceptibility~\cite{chen2024susceptible} and persuasive capabilities of LLMs~\cite{breum2024persuasive, salvi2024conversational}, showing that contextual information in specific domains can enhance LLMs' persuasiveness~\cite{matz2024potential}. In particular, LLMs have shown competence in generating persuasive messages in public health~\cite{karinshak2023working} and personalized advertising~\cite{meguellati2024good}.

This  body of work  focuses on the capabilities of LLMs in one-on-one interactions within isolated contexts. However, their potential within broader interconnected structures, such as their ability to create information cascades among the users of
a social network, remains largely unexplored. Such capacity can have significant implications, including fostering discussions, shaping opinions, and driving behavioral changes among users. In light of this, we address the following research question: \emph{Can LLMs learn to generate meaningful content that maximizes user engagement on social networks?}

\begin{figure}[t!]
    \centering
    \includegraphics[width=0.95\columnwidth]{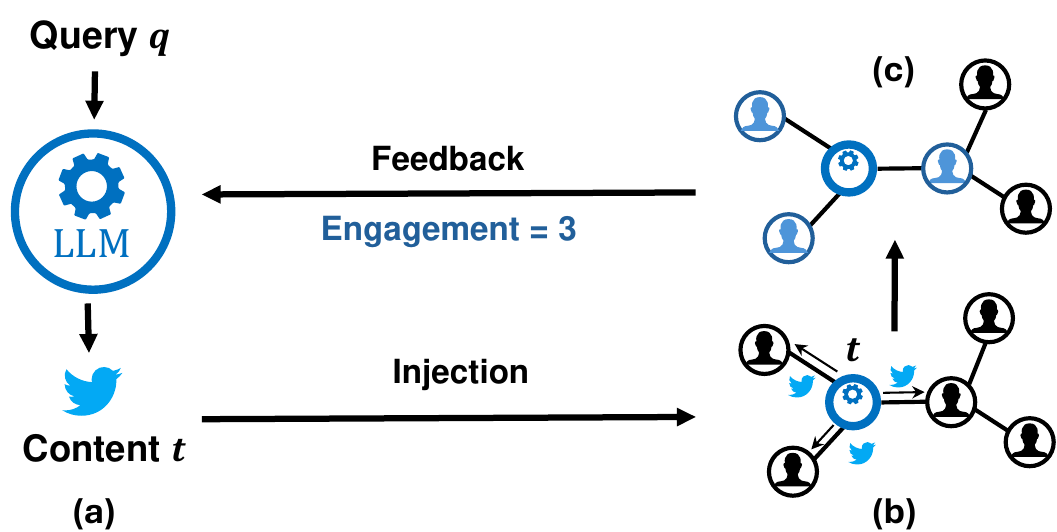}
    \caption{
        A visual representation of the proposed framework:
        (a) A query \( q \) is prompted to the LLM, which generates content \content.
        (b) The content \content is injected into the social network (i.e., posted by a node that corresponds to the LLM agent) and starts to propagate into the network according to an engagement model.
        (c) The number of active users at the end of the propagation is returned to the LLM as the observed reward.
        The process repeats until convergence.
        }
    \label{fig:framework}
\end{figure}

Our proposal to answer this question is a framework based on reinforcement learning, whose high-level vision is depicted in Figure~\ref{fig:framework}.
We begin by prompting the LLM with a query $q$ describing the topic of interest. The LLM generates a text $t$, which is then injected into the social network by an agent (a node of the network), acting as the injection point and representing the LLM itself. The content $t$ propagates through the network, generating a certain level of engagement, i.e., the number of users interacting with $t$. This observed engagement, combined with a \emph{fluency score} of the text $t$, is returned to the LLM as a reward
which is used to fine-tune the LLM's generation abilities to maximize the expected reward. This process is repeated until convergence.

The framework presented in Figure~\ref{fig:framework} can be realized by means of \emph{Reinforcement Learning with Human Feedback} (RLHF), an approach which has gained popularity due to its effectiveness in various learning tasks~\cite{kaufmann2023survey}, including fine-tuning of LLMs~\cite{10.5555/3600270.3602281}.
However, conducting actual live experiments involving propagating content through a social network and waiting for responses, is complex and time consuming.
This is a common challenge in RLHF frameworks~\cite{casper2024open}. To overcome this, we leverage an approach based on \textit{Reinforcement Learning from Simulated Feedback} (RLSF). Specifically, we adopt a proxy model to simulate the network's response, employing a formal engagement mechanism, which borrows ideas from both information propagation and opinion dynamics literature~\cite{deffuant2000mixing}.
In addition to bypassing the temporal cost and complexity of live experiments, thus enabling an efficient and effective feedback loop between the LLM agent and the network under study, such an approach also allows to control over endogenous factors such as the LLM's position within the social network and the distribution of opinions on a given topic. Thanks to the feedback loop, the LLM learns to adapt to the current topic, the social network structure, and the distribution of opinions in the network. For instance, if the opinion of the network on the given topic is prevalently negative, the generated content should have a negative leaning so to engage as many users as possible. 

Using our framework, we analyze the capabilities and limitations of LLMs in tackling the given task, specifically considering the relative positions of the LLM as an agent within the social network and the distribution of opinions in the network on the given topic. The experimental evaluation suggests that our framework is \textit{agnostic} to the underlying propagation model and fully \textit{adaptive}, meaning that the LLM agent automatically adapts the sentiment of the generated content to the underlying opinion distributions. We further show that our fine-tuning process is effective over both synthetic and real social networks, and that the engagement produced with generated posts by means of our propagation model, is comparable to that of actual content within a real network.


The rest of the paper is structured as follows. Section~\ref{sec:background} reports an overview of the literature related to our problem. We formalize the problem in Section~\ref{sec:model} and provide the technical details of the approach in Section~\ref{sec:algo}. 
Section~\ref{sec:eval} and \ref{sec:findings} outline the experimental design used to evaluate the proposed methodology and analyze how the experimental results address the formulated research questions.
Finally, in Section~\ref{sec:conc}, we discuss the implications of our findings, along with potential extensions and limitations of the study.

\section{Background and Related Work}\label{sec:background}

In the previous section, we have already covered the literature on the persuasion capabilities of LLMs in one-to-one interactions. We next review the rest of the related literature.



\spara{LLM-Agents in complex environments.} The integration of Agent-Based Models (ABMs) and Large Language Models has been an emerging area of research.
Notable studies, such as those by \citet{vezhnevets2023generative} and \citet{gurcan2024llm}, focus on utilizing LLMs as social agents within ABMs to simulate micro-level behaviors and observe macro-level patterns. These micro-level behaviors can correspond to opinion dynamics models~\cite{chuang2024simulating}, and the resulting macro-level patterns can lead to known universal laws such as scale-free networks~\cite{de2023emergence}. Recent studies have started to place LLM agents into more complex environments to assess their ability to mimic human behavior~\cite{park2023generative}, and to understand their capabilities in negotiation~\cite{bianchiwell}, code development~\cite{qian2024chatdev}, and even in tasks like building houses~\cite{chen2023agentverse}. A concurrent line of inquiry is whether LLMs can learn to align their behavior with social norms~\cite{frankensocial}.

Nevertheless, the use of LLMs in large-scale social media environments, where word-of-mouth dynamics and complex interactions are crucial, has not been adequately explored, overlooking cascading effects that extend beyond one-to-one interactions.

\spara{Social media environments.}
The adoption of information diffusion models in data mining methodologies has been thoroughly investigated for designing viral marketing strategies~\cite{kempe2003maximizing, leskovec2007dynamics, chen2010scalable, lu2013bang, Barbieri2013c, barbieri2014influence, tu2022viral}. Various optimization problems have been tackled such as selecting influential seeds~\cite{kempe2003maximizing, tu2022viral}, designing product features~\cite{barbieri2014influence},
or other variations~\cite{lu2013bang, aslay2015viral}.
Yet, a crucial practical dimension remains largely unexamined: how to compose textual content that effectively propagates through a network.
In this work, we bridge this gap by proposing a methodology that enables LLMs to generate content maximizing engagement on social platforms.


\spara{Proxy for human feedback.}
Regardless of the type of objectives that machine learning models have, a correct reward function is necessary.
In particular, the challenges associated with human feedback in evaluating LLMs have led to alternative approaches, such as LLM-as-a-Judge~\cite{huang2024empirical, verga2024replacing} and human-in-the-loop methods~\cite{boubdir2023prompts, amirizaniani2024developing}, which aim to enhance scalability and efficiency. While our work shares the goal of reducing the reliance on human experts for data annotation, it introduces a fundamentally novel approach by simulating context-dependent feedback that incorporates the social network structure.
Hence, this innovation also minimizes the need for slow, costly, and potentially unethical online experiments, offering a more efficient and ethically sound alternative.
To the best of our knowledge, no existing work leverages simulated social media engagement models as feedback mechanisms to fine-tune LLMs, addressing a critical gap in current research.

\section{Engagement Model}\label{sec:model}


As discussed in Section \ref{sec:intro}, we adopt a \textit{Reinforcement Learning from Simulated Feedback} approach. This method not only helps us avoiding the complexity of live experiments but also allows us to control contextual settings, such as the distribution of opinions and the LLM’s position within the social network.
Given our goal to study the capabilities of LLMs in generating content that resonates across the network, we propose a novel engagement model which borrows
from information propagation~\cite{kempe2003maximizing} the cascade mechanism, and from
opinion dynamics~\cite{deffuant2000mixing}, the single engagement mechanism.

More formally, we consider a social network represented as a
directed graph $\network = (\mathcal{V},\mathcal{E})$ where nodes are users and a directed edge $(u, v) \in \mathcal{E} \subseteq \mathcal{V} \times \mathcal{V}$ indicates that $v$ is a ``follower'' of $u$: thus $u$ can propagate content to $v$, but not necessarily vice versa. We are also given a vector $\vec{x} \in [0,1]^{|\mathcal{V}|}$, associating each node $u \in \mathcal{V}$ with a
 scalar $x_u$ between 0 and 1, representing its opinion on a given topic.
We interpret $x_u \approx 1$ as being in favor of the topic, while $x_u \approx 0$ indicates disapproval of the topic.

Consistently with the \emph{Independent Cascade} model~\cite{kempe2003maximizing},
the propagation of a piece of content \content is modeled in discrete timesteps and independently of other content that might be propagating concurrently. More precisely,
a new content \content, having a leaning  $s_\content \in [0,1]$ w.r.t. a given topic, is injected in the network (i.e., posted by a node $u \in \mathcal{V}$) at time step $T_0$. When a user posts a content, their followers become aware of the content and might engage with it. If they do, they become ``active'' on \content at the subsequent timestep and their followers, in turn, become aware of $t$. The decision of engaging with the content is modeled according to the \emph{Bounded Confidence Model}~\cite{deffuant2000mixing}, which is one of the most studied models in the opinion dynamics literature: a node $u \in V$ engages with the content $t$, to which it is exposed, \emph{iff} $|s_t - x_u| < \epsilon$, where $\epsilon \in [0,1]$ is the given confidence bound.
This process repeats until no new nodes become active, with each node activating at most once on the specific content $\content$. 

\newcommand\mycommfont[1]{\textcolor{blue}{#1}}
\SetCommentSty{mycommfont}
The pseudo-code of our engagement model is provided in  Algorithm~\ref{alg:engagement-model}.
\begin{algorithm}
\small
\caption{Engagement Model $\mathcal{M}_{\epsilon}$}
\label{alg:engagement-model}
\SetKwInOut{Input}{Input}
\SetKwInOut{Output}{Output}
\SetKw{Return}{return}
\Input{Broadcasting node sentiment $s_\content$; broadcasting node $u \in \mathcal{V}$; Network \network and node opinions $\vec{x}$; bounded confidence threshold $\epsilon$}
\Output{Cardinality of the set of engaged users}
$\mathcal{A} = \mathcal{C} = \{ u\}$ \tcp*{Init sets of active and contagious users}
\While{$\mathcal{C} \neq \emptyset$}{
   $\mathcal{C}' = \emptyset$ \tcp*{Init new set of contagious users}
   \ForEach{$v \in \mathcal{C}$}{%
       \ForEach{$w \in \mathcal{N}_v$}{
           \If{$w \notin \mathcal{A}$ \textbf{and} $\lvert s_\content - x_w \rvert \leq \epsilon$}{
               $\mathcal{A} = \mathcal{A} \cup \{w\}$ \tcp*{$w$ becomes active}
               $\mathcal{C}' = \mathcal{C}' \cup \{w\}$ \tcp*{$w$ becomes contagious}
           }
       }
   }
   $\mathcal{C} = \mathcal{C}'$ \tcp*{Update set of contagious users}
}
\Return{$|\mathcal{A}|$}
\end{algorithm}


\spara{Engagement maximization.} The goal of our study is to teach the LLM to generate content $t$ that maximizes the total engagement (i.e., the number of users that activate on $t$) in the social network under the engagement model $\mathcal{M}$ described above. We are given the following elements: 
\begin{enumerate}[leftmargin=*]
    \item The social network structure $\network = (\mathcal{V},\mathcal{E})$  and a designated node  $u_L \in \mathcal{V}$  as the \emph{content writer}, serving as the point of injection of \content in $\network$.
    \item The underlying topic, provided to the LLM by means of a query, or prompt $q$,  and the vector of nodes opinion $\vec{x} \in [0,1]^{|\mathcal{V}|}$ regarding the topic.
    \item Two black-box functions, $\utilitymodel$ and $\mathcal{F}$, which, given a content $\content$, provide \begin{enumerate*}[label=(\roman*)] \item the leaning $s_\content$ of $t$ with respect to  the given topic; \item A \emph{fluency score} $f_\content$ representing the readability of $t$ in natural language\end{enumerate*}.
    \item An engagement model $\propagationprotocol_\epsilon$, which determines content propagation based on an associated content value and a bounded confidence level $\epsilon \in [0,1]$.
\end{enumerate} 
Our objective is then to have the LLM generate content $t$ that \begin{enumerate*}[label=(\roman*)]
\item exhibits high fluency score $f_t$, and 
\item maximizes engagement within the network, i.e., activates the largest possible number of nodes
\end{enumerate*}. Formally, given the set $\mathcal{T}$ of all possible texts that achieve an adequate level of fluency, we aim at finding $t^\ast = \arg \max_{t \in \mathcal{T}} \mathcal{M}_\epsilon(s_t, u_L|\network, \vec{x})$. 

In the next section, we propose a method to tackle this problem using reinforcement learning.

\label{sec:engagement-model}

\section{Proposed Method}\label{sec:algo}
In this section, we provide a detailed description of the individual components of our framework: Large Language Models and the Reinforcement Learning technique.



\spara{Preliminaries.}
A Large Language Model (LLM) can be formally described as a function $\mathrm{LLM}_{\theta}(x)=y$ that stochastically maps input sequences of tokens $x = [x_1, x_2, \dots, x_n]$ to an output sequence $y = [y_1, y_2, \dots, y_m]$, where $n$, $m$ represent the length of the input and the output sequence, respectively.
The model defines $P_\theta(y|x)$, the probability distribution of $y$ given $x$, capturing complex patterns and relationships in natural language between the two sequences.
The response $y$ is then sampled from the distribution $P_\theta(\cdot|x)$.

For our purposes, we fine-tune the underlying LLM agent using a Reinforcement Learning (RL) mechanism (widely adopted for optimizing LLMs~\cite{10.5555/3600270.3602281}).
Generally, RL is suitable for scenarios where an agent needs to learn to interact within a dynamic environment.
The agent's goal is to learn a \textit{policy}—a strategy or behavioral pattern—through feedback it receives from the environment, in the form of rewards or punishments.

In our framework,
given a prompt $x$, we generate a response $y \sim P_\theta(\cdot|x)$ and compute a reward $\finalreward(y)$ that scores the response on the specific application domain.
This reward can hence be used within a Policy-gradient strategy aimed at optimizing
$$
\mathcal{L}(\theta) = \mathbb{E}_{y\sim P_\theta(\cdot|x)} \left[ \finalreward(y) - \beta   \log \frac{P_\theta(y|x)}{P_{\theta'}(y|x)}\right],
$$
where $\theta'$ represents the reference (not fine-tuned) model, while the second term incorporates the KL-divergence between the outputs of the fine-tuned model and the reference model, serving as an additional signal to ensure that the generated responses do not deviate significantly from the reference model. Here $\beta$ is the coefficient that controls the weight of this KL divergence penalty: in our experiments we keep the default\footnote{\url{https://github.com/huggingface/trl/blob/main/trl/trainer/ppo_config.py}} value 0.05.

Researchers have refined the above optimization problem to avoid learning instability.
In particular, the Proximal Policy Optimization (PPO)~\cite{schulman2017proximal} strategy, which we adopt in our framework, introduces constraints on policy updates through a clipped objective function.
This ensures that updates do not destabilize the learning process by limiting the extent of policy changes, thereby providing stability and efficiency in fine-tuning large language models.

\spara{Fine-tuning framework.}
Algorithm~\ref{alg:procedure} presents, in pseudocode, the basic flow of our proposed LLM fine-tuning framework based on reinforcement learning, consisting of the following steps:

\begin{algorithm}[t!]
\caption{Fine-tuning framework}
\small
\label{alg:procedure}
\SetKwInOut{Input}{Input}
\SetKwInOut{Output}{Output}
\Input{Language Model \llm; network $\network = (\mathcal{V},\mathcal{E})$ with opinions $\vec{x}$; injection point $\pos \in \mathcal{V}$; prompt query  \prompt; Engagement model $\mathcal{M}_{\epsilon}$; black-box functions \utilitymodel and $\mathcal{F}$; max number of iterations \maxiterations; KL-Divergence threshold \klthreshold}
\Output{Fine-tuned \llmm{\theta^*}}

\smallskip

$j = 0$; $\kappa = 0$; $\theta^{(0)} = \theta$\\
\While{$j < \maxiterations$ \textbf{and} $\kappa < \klthreshold$}{
    $\content \gets \llmm{\theta^{(j)}}(\prompt)$;    $\kappa\gets \mathbb{E}_{\content\sim P_{\theta^{(j)}}(\cdot|q)}\left[\log\frac{P_{\theta^{(j)}}(\content|q)}{P_{\theta^{(0)}}(\content|q)}\right]$\\
    $\opinion_\content \gets \utilitymodel(\content)$;  $\readability_\content \gets \mathcal{F}(\content)$ \\
    $\mathcal{A} \gets \propagationprotocol_{\epsilon}(\opinion_\content, \pos|\network,\vec{x})$;
    $\finalreward \leftarrow \left(\readability_\content \cdot |\mathcal{A}|\right)^{1/2}$ \\
    Compute $\theta^{(j+1)}$ by updating $\theta^{(j)}$ and using \finalreward as reward\\
    $j \gets j + 1$
}
\end{algorithm}


\begin{enumerate}[leftmargin=*]
    \item \underline{\emph{Content generation.}} Starting from \prompt, the LLM generates the content $\content = \llm(\prompt)$ (line 3). The the KL-divergence between the outputs of the fine-tuned model and the reference model is also computed.

    \item \underline{\emph{Sentiment inference.}} We use a pre-trained model \utilitymodel to compute a sentiment value $\opinion_\content$ associated with the content \content, i.e., $\utilitymodel(\content) = \opinion_\content$ (line 4). This utility model is a plug-and-play component that can be adapted to different tasks. In our experiments, we adopt a DistilBERT model~\cite{Sanh2019DistilBERTAD} pretrained for sentiment classification as our function \utilitymodel.

    \item \underline{\textit{Fluency score.}} To ensure that \content conveys proper semantics, we compute a readability value associated with its content. While perplexity is a common measure for evaluating the generation quality of LLMs~\cite{gonen-etal-2023-demystifying}, it is sensitive to vocabulary size and sentence length. Studies~\cite{M+20} highlight its low correlation with readability. \citet{10.1162/coli_a_00398}, study the adequacy of several readability scores and assess their performance on several datasets. For our purposes, we adopt as our fluency function $\mathcal{F}$ the Flesch–Kincaid (FK)~\cite{Kincaid1975DerivationON} formula (line 4).
    This score quantifies the grade level required to understand an English statement: the higher the value, the higher its complexity and syntactic quality.
    It is a crucial component of the final reward \finalreward, as it ensures the LLM generates meaningful and fluent statements.

    \item \underline{\textit{Propagation simulation.}} The LLM, corresponding to the node $\pos$, posts $t$ on the social network \network. The content $t$ propagates according to $\mathcal{M}_{\epsilon}$. The outcome of the simulation is  the set of active users, $\mathcal{A}$ (line 5). The size of this set, $|\mathcal{A}|$, is the main component of the reward \finalreward, as it quantifies the engagement produced by \content.

    \item \underline{\textit{Reward computation.}}  We compute \finalreward as the geometric mean of readability score and the content's virality in the network, i.e., $\finalreward = \left(\readability_\content \cdot |\mathcal{A}|\right)^{1/2}$ (line 5).
    \item \underline{\textit{Policy update.}} We use $\finalreward$ as the reward in the current step of the RL training procedure (line 6) and repeat the process (lines 3-7).
\end{enumerate}

The fine-tuning procedure stops if either of the following conditions occurs: (\textit{i}) a (fixed) number of iterations \maxiterations is reached, or (\textit{ii}) the KL-divergence exceeds a given threshold \klthreshold.  

An interesting feature of our approach is that, although the reward computation (line 5) scales with the number of nodes, the primary computational cost lies in the LLM backpropagation (line 6) which remains independent of the network size.

\section{Experiments Setup}\label{sec:eval}
We evaluate the proposed framework through a series of experiments aimed at answering the following research questions:
\begin{itemize}
\item[\textbf{RQ1:}] \textit{Can the LLM agent learn to generate content that maximizes engagement? How do different network conditions (modularity, homophily, opinions distribution)  affect the performance?}
\item[\textbf{RQ2:}] \textit{Are the generated contents realistic and meaningful compared to real content propagated on a social platform?}
\item[\textbf{RQ3:}] \textit{How does the proposed framework compare to other content generation methods from the literature?}
\end{itemize}
We address RQ1 on synthetically generated networks with a grid of configurations, RQ2 using real-world data obtained from $\mathbb{X}$ social media, and RQ3 by comparing our approach to baselines and state-of-the-art language models on both synthetic and real-world data.

\spara{Synthetic data.}
To rigorously evaluate our proposed methodology across diverse initial conditions, we devise a synthetic data generator, based on random network model inspired by~\cite{cinus2022effect}, and capable of producing realistic networks while allowing to control the key parameters influencing information diffusion, such as homophily, modularity, and opinion distribution. The generator takes as input five parameters  -- homophily level $\eta$, network modularity $\mu$, shape parameters $\alpha$ and $\beta$ for opinion distribution, the number of nodes $N$ -- and outputs a directed graph $\mathcal{G}=(\mathcal{V}, \mathcal{E})$, also assigning an opinion value $x_u \in [0,1]$ to each node $u \in \mathcal{V}$. The generation process starts by constructing the set of edges $\mathcal{E}$ and assigning nodes to non-overlapping communities $c: \mathcal{V} \rightarrow C$ using the LFR model~\cite{lancichinetti2008benchmark}, known for producing community sizes that adhere to a realistic power-law distribution. The modularity parameter $\mu$ is an input of the LFR model and controls the community segregation within the network. Each community $c_i \in C$ gets assigned an initial opinion drawn from a Beta distribution\footnote{Beta distribution has been chosen for its flexibility to accommodate various distribution shapes such as uniform, unimodal, and bimodal.} $o_{c_i} \sim Beta(\alpha,\beta)$. For each node $u \in \mathcal{V}$, its opinion $x_u$ is determined by a Bernoulli trial with probability $\eta$ (i.e., homophily). If successful, the node adopts the opinion of its community $x_u=o_{c(u)}$; otherwise, it independently samples an opinion from the same Beta distribution.

This approach allows us to generate synthetic networks where \emph{initial homophily}, \emph{modularity}, and \emph{opinion distribution} can be controlled.
In our experiments, we fix the opinion distribution to be either \textbf{positive} (skewed on 1), \textbf{negative} (skewed on 0), \textbf{neutral} (centered on 0.5), or \textbf{uniform} (uniformly distributed on [0, 1]).
Further, we force homophily to be either \textbf{low} (0.25) or \textbf{high} (0.75). Similarly, we tuned the modularity as either \textbf{low} or \textbf{high}. 

\begin{figure*}[t!]
    \centering
    \begin{tabular}{ccccc}
        \includegraphics[width=0.185\textwidth]{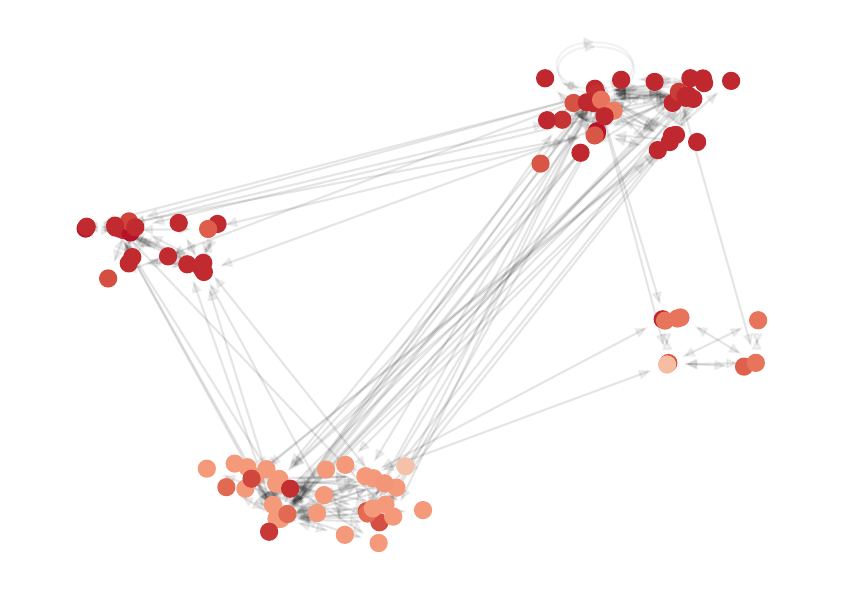} &
        \includegraphics[width=0.185\textwidth]{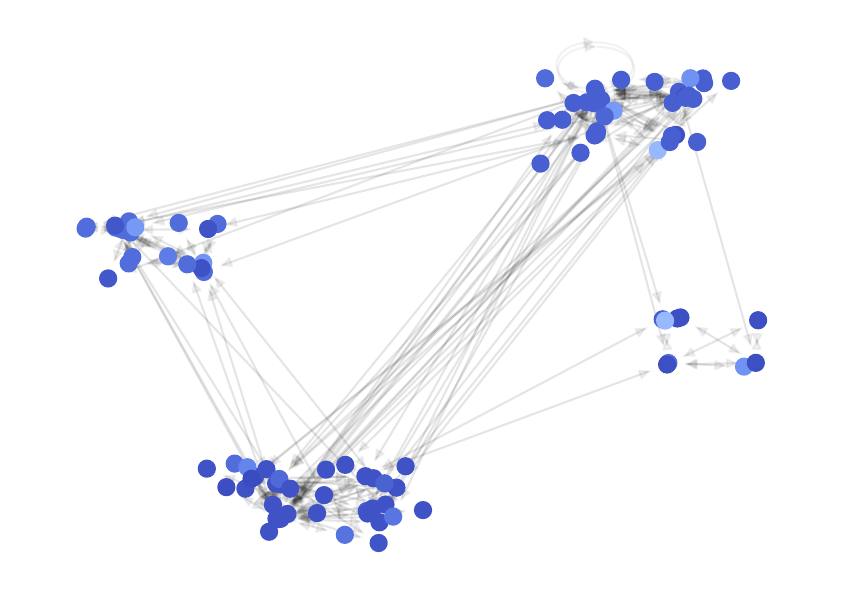} &
        \includegraphics[width=0.185\textwidth]{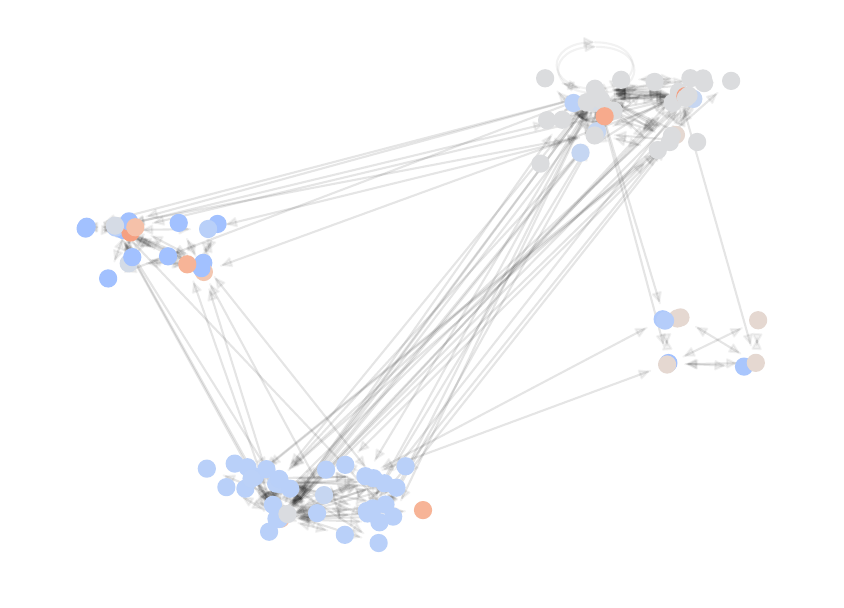} & 
        \includegraphics[width=0.185\textwidth]{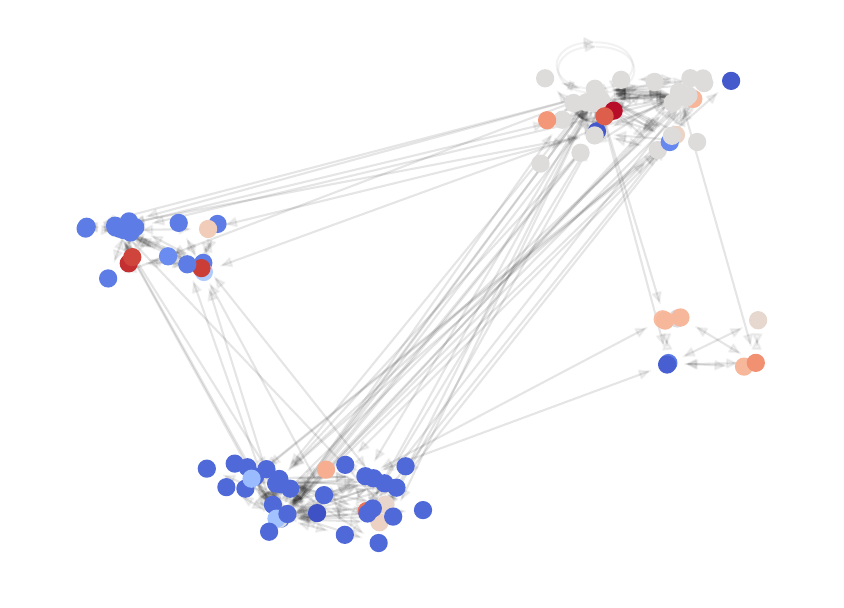} &
        \includegraphics[width=0.185\textwidth]{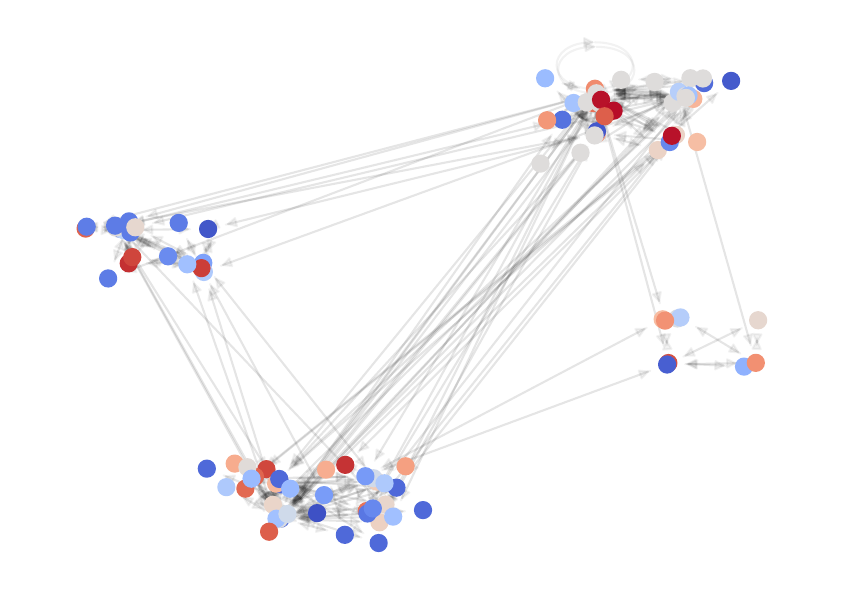} \\
    \end{tabular}
    \centering
    \begin{minipage}[t]{0.2\textwidth}
        \includegraphics[width=\textwidth]{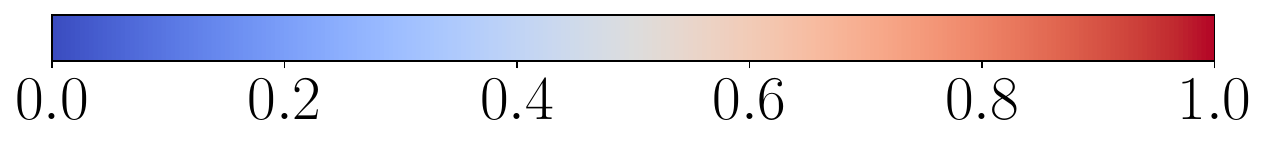}
    \end{minipage}
    \caption{
        Different opinion settings (from left to right): positive/negative/neutral/uniform/uniform. All networks except for the last one exhibit high modularity and high homophily (i.e., neighbors have similar opinions). The last graph is indeed an example of network with high modularity but low homophily (i.e., neighbors' opinions are heterogeneous). Colors represent node opinions, whose values are denoted by the color bar. 
    }
    \label{fig:synthetic-graphs-with-opinions}
\end{figure*}
\begin{figure}[h!]
    \centering
    \begin{tabular}{cc}
        \includegraphics[width=0.4\columnwidth] {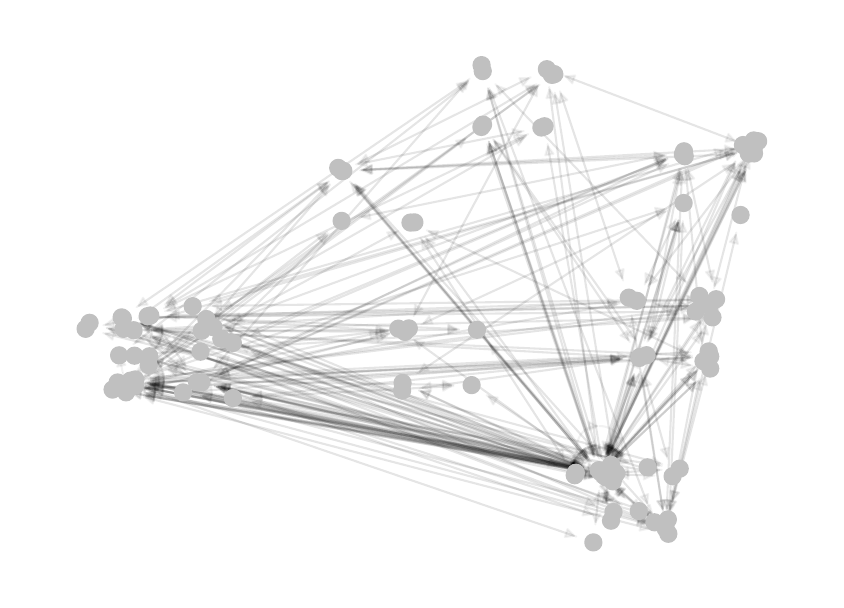} &
        \includegraphics[width=0.4\columnwidth]{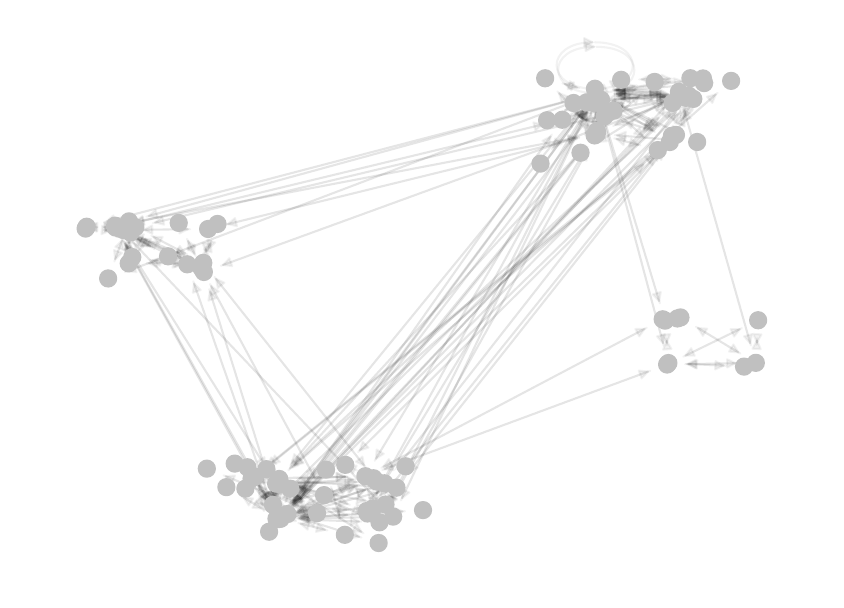} 
    \end{tabular}
    \caption{
        Networks with low modularity (left) and high modularity (right). The former presents high connectivity among the nodes, while the latter exhibits clustered communities.
    }
    \label{fig:synthetic-modularity}
\end{figure}


Figures~\ref{fig:synthetic-graphs-with-opinions} and~\ref{fig:synthetic-modularity} show, respectively, the different opinion distributions across network structures with high/low homophily, and modularity variations.
 Taking into account the network characteristics, we consider the following configurations for the \emph{position of the LLM in the network}:
\begin{itemize}[leftmargin=*]
    \item \textbf{Echo-low (-high)}: the LLM agent is placed within the echo-chamber with the lowest (resp. highest) average opinion.
    \item \textbf{Comm-largest (-smallest)}: we locate the LLM inside the largest (resp. smallest) community.
    \item \textbf{Central}: the LLM is the node with the highest betweenness centrality in the graph.
\end{itemize}

\begin{figure}[t!]
    \centering
    \begin{subfigure}[b]{0.49\columnwidth}
        \includegraphics[width=\columnwidth]{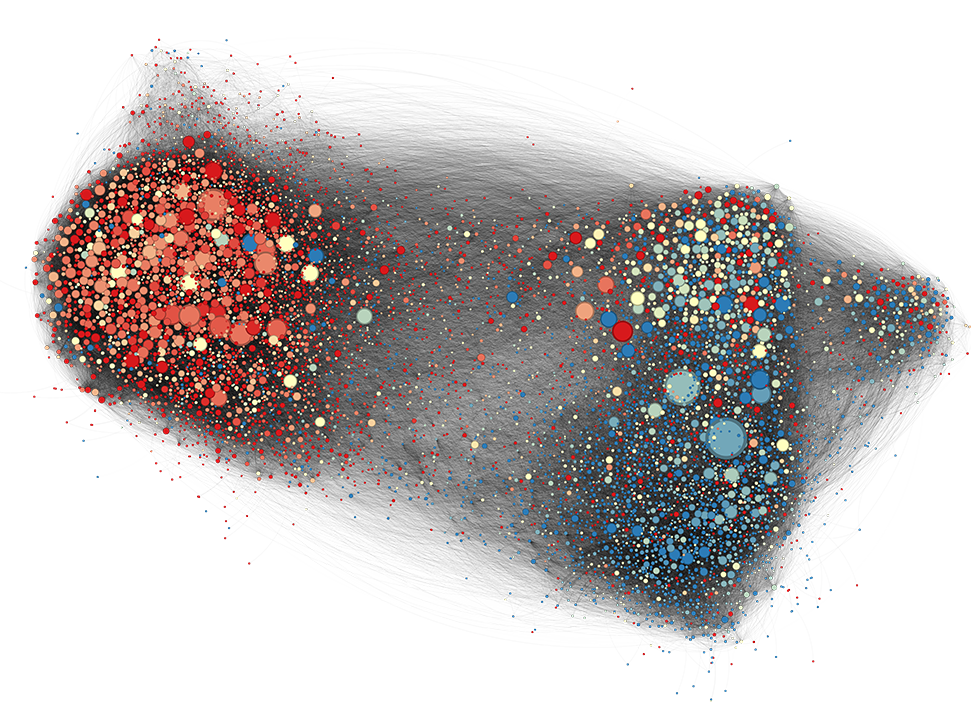}
        \caption{\textsl{Brexit} dataset.}
        \label{fig:brexit-viz}
    \end{subfigure}
        \begin{subfigure}[b]{0.49\columnwidth}
        \includegraphics[width=\columnwidth]{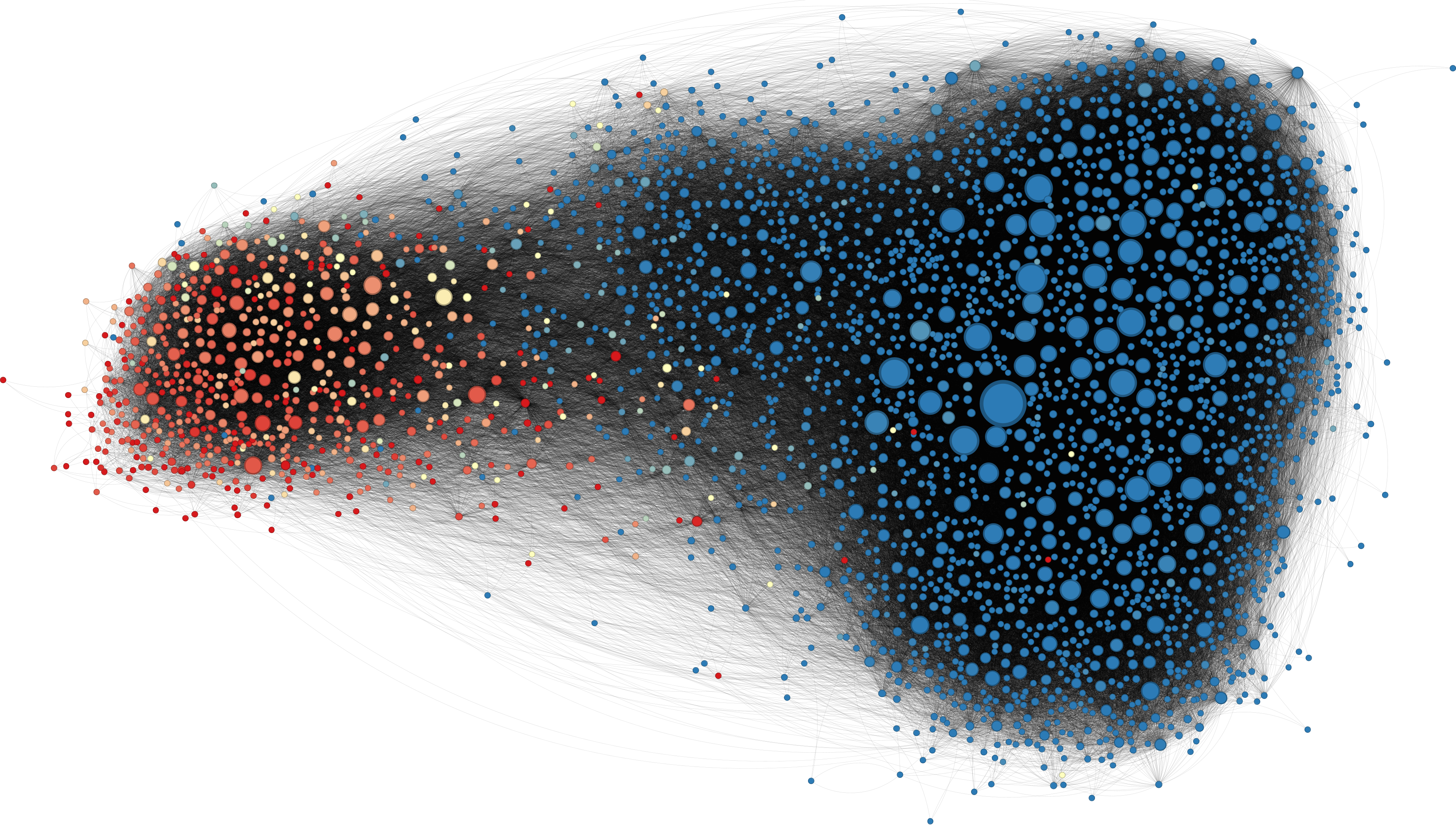} 
        \caption{\textsl{Italian Referendum} dataset.}
        \label{fig:referendum-viz}
    \end{subfigure}
    \caption{Visualization of the $\mathbb{X}$ follow networks. Nodes are colored on a scale from blue (0: "Remain"/"No") to red (1: "Leave"/"Yes"), and their size resembles their out-degree.}
    \label{fig:real_networks}
\end{figure}
\spara{Real data.}
We also consider two real-world datasets containing content about a controversial topics propagated over $\mathbb{X}$ social network. The first dataset, \textsl{Brexit},
focuses on the remain-leave discourse before the 2016 UK Referendum on exiting the EU~\cite{ZHU2020102031}. We use the version in~\cite{minici2022cascade} consisting of 7,589 users, 532,459 directed follow relationships, and 19,963 tweets, each associated with a binary stance.
The dataset is preprocessed according to~\cite{minici2022cascade} to assign each user a scalar value  $x_u \in [0,1]$, referred to as \textit{opinion}, representing the average stance of the tweets retweeted by user $u$.
The stance of each tweet is either 0 (``Remain'') or 1 (``Leave'').
The second dataset, \textsl{Italian Referendum}, was gathered during the Italian constitutional referendum in 2016~\cite{lai2018stance} and processed according to~\cite{minici2022cascade}. This dataset includes 2,894 users, 161,888 edges, and 41,001 tweets. Each user is associated with a scalar value  $x_u \in [0,1]$ , corresponding to their stance on the referendum question, where values represent support for ``No'' or ``Yes''.
The structure of the networks and users' opinions are depicted in Figure~\ref{fig:real_networks}.
Both datasets exhibit two separated, homophilic, and polarized communities, and a smaller, more heterophilic mixed-opinions community.

\spara{Settings.}
Although other modeling choices are possible\footnote{For instance, the \textit{Generation from Scratch} task: i.e., ``\textit{Generate a post about} [TOPIC]''.}, in our experiments we focus on a \emph{Query Completion Task}, as we found that Query Completion achieves faster convergence. We defer a detailed study on the effects of task specification to future work. Specifically, in the synthetic setting, we use the query ($q =\ $``\textit{Cats are the most...}''), while in the real experiments we adopt $q=\ $``[\textit{Brexit}]/[\textit{2016 Italian constitutional referendum}] \textit{is the most...}''. Given the query $q$, we obtain \content by concatenating \prompt and the generated text $\generated=\llm(q)$, i.e., $\content = \prompt \mathbin\Vert \generated$. To compute the opinion value $\opinion_\content$, we adopt a DistilBERT model~\cite{Sanh2019DistilBERTAD}, pretrained for sentiment classification. 

Concerning the LLM agent, we adopt the 2B version of Gemma
\cite{gemma}, a lightweight models family released by Google and based on Gemini\footnote{\url{https://deepmind.google/technologies/gemini/}} technology. Despite its smaller size, Gemma outperforms other models of similar size in tasks such as understanding, reasoning, and safety. We fine-tune the model exploiting the \texttt{PPOTrainer}\footnote{\url{https://huggingface.co/docs/trl/main/en/ppo_trainer\#trl.PPOTrainer}} class from the \texttt{trl} package, which enables training language models with custom rewards. The maximum number of training steps is set to $80$ for the synthetic data and $500$ for the real-world datasets, while the threshold $\tau$ for monitoring the KL-divergence to $75$. We also conducted experiments with other state-of-the-art lightweight LLMs, such as Mistral-7B~\cite{mistral},
LLaMA2-7B~\cite{llama2},
and GPT-2~\cite{gpt2}.
However, these experiments demonstrated that Gemma-2B provides the best trade-off between performance and computational efficiency.
Finally, for the engagement model, we set $\epsilon=0.2$.


\begin{figure*}[t!]
    \centering
    \begin{minipage}[t]{\linewidth}
    \centering
       \includegraphics[width=0.8\linewidth]{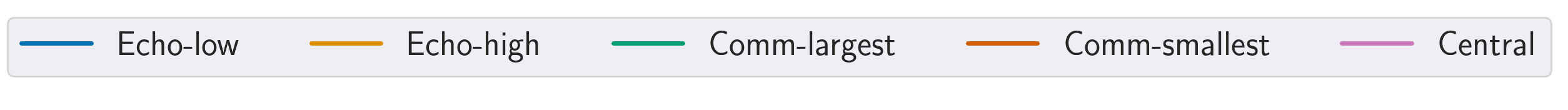}
    \end{minipage}
    \begin{tabular}{cccc}
        \includegraphics[width=0.22\textwidth]{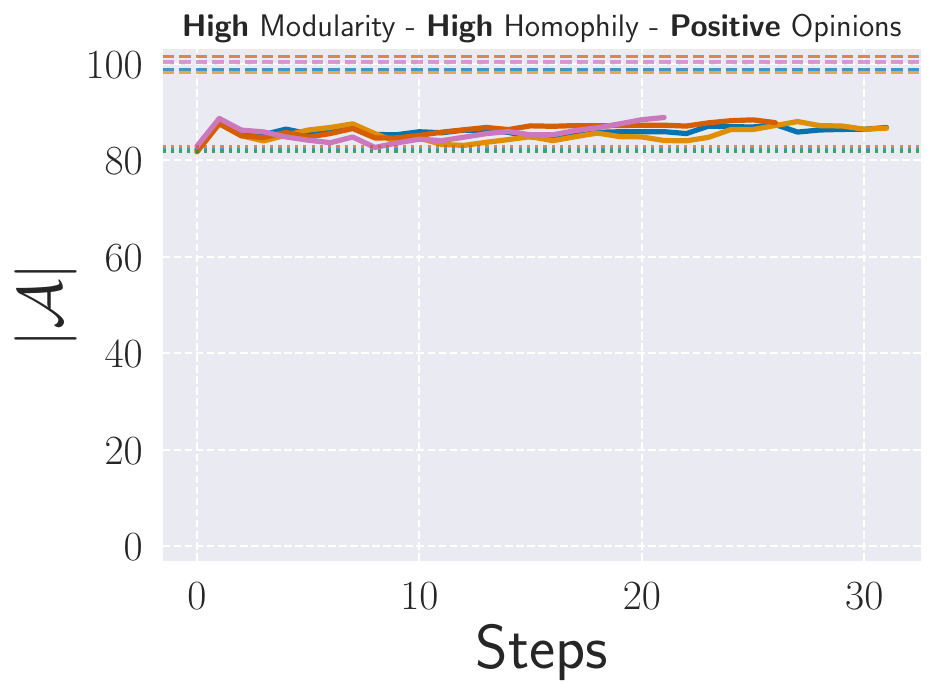} &
        \includegraphics[width=0.22\textwidth]{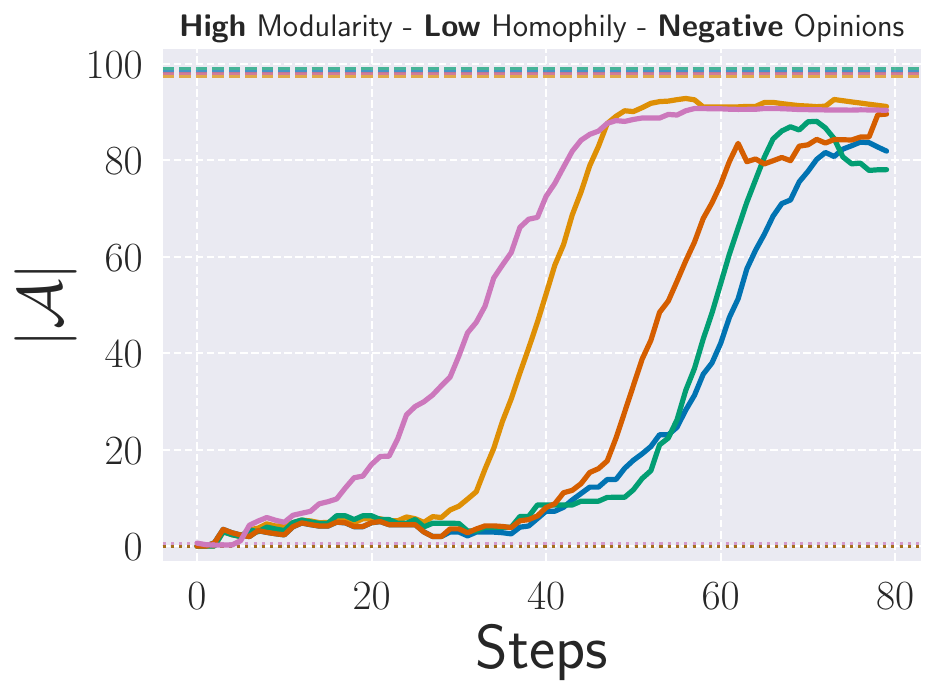} &

        \includegraphics[width=0.22\textwidth]{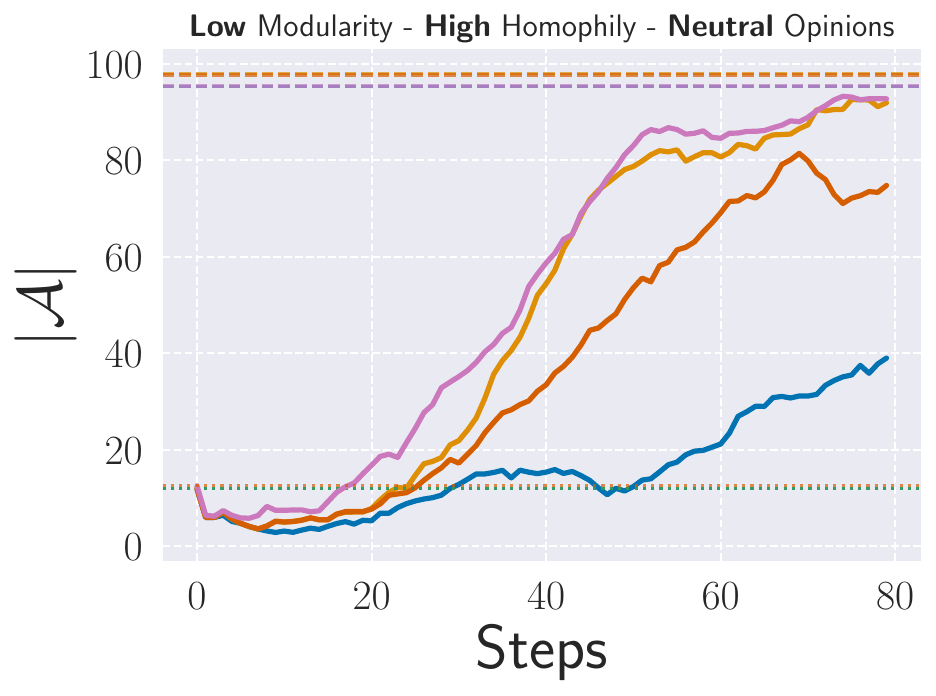} &
        \includegraphics[width=0.22\textwidth]{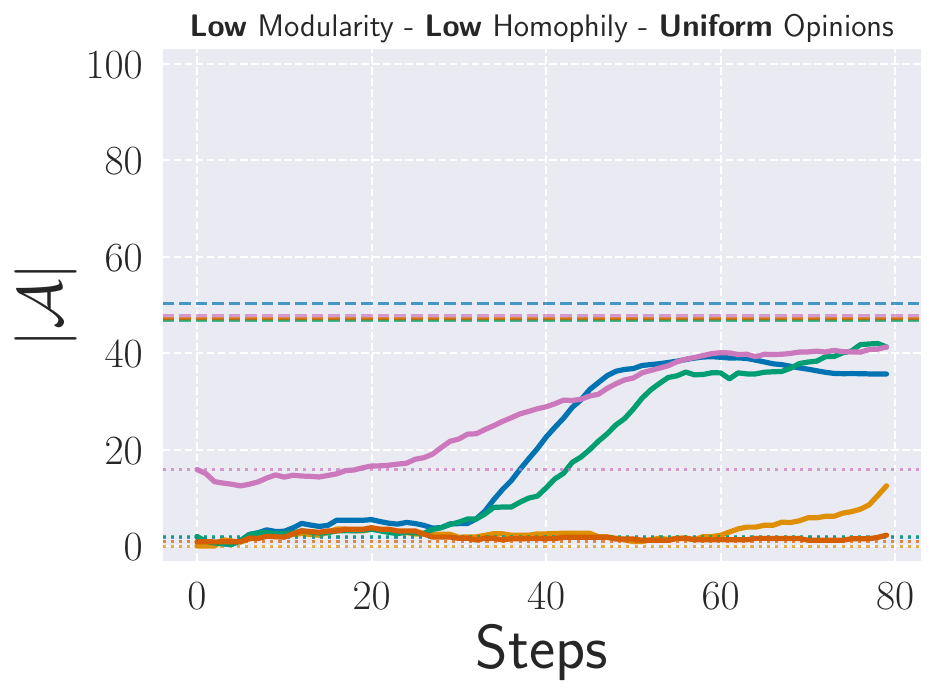} \\
    \end{tabular}
    \vspace{-4mm}
    \caption{Engagement $|\mathcal{A}|$ at each step of our fine-tuning procedure. Columns refer to different opinion distributions: positive/negative/neutral/uniform (left to right). Each plot depicts the trend varying the network structure in terms of modularity and homophily. Colors indicate different positions of the LLM agent. Dashed lines represent the maximum engagement within that configuration, whereas dotted lines indicate its lower-bound (i.e., $|\mathcal{A}|$ produced by the non-finetuned LLM).}
    \label{fig:synthetic_engagement}
\end{figure*}


\begin{figure*}[ht!]
    \centering
    \begin{tabular}{cccc}
        \includegraphics[width=0.22\textwidth]{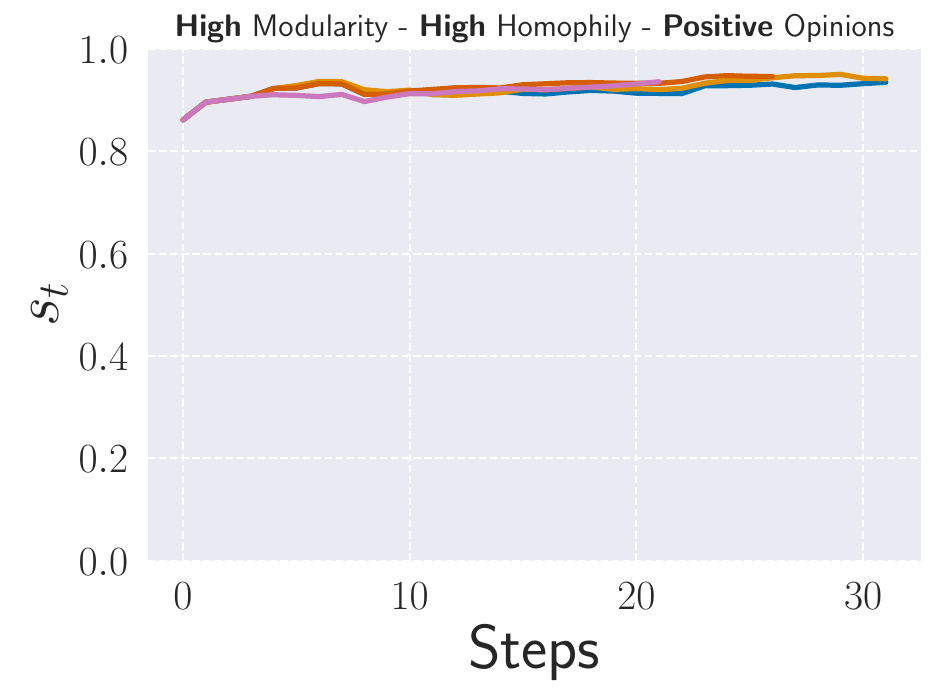} &
        \includegraphics[width=0.22\textwidth]{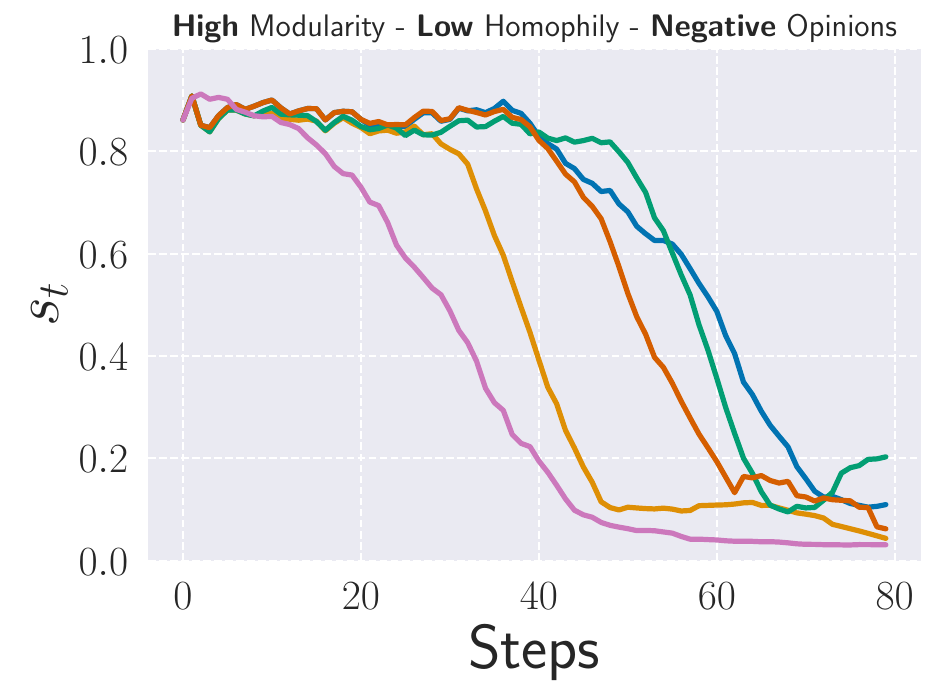} &

        \includegraphics[width=0.22\textwidth]{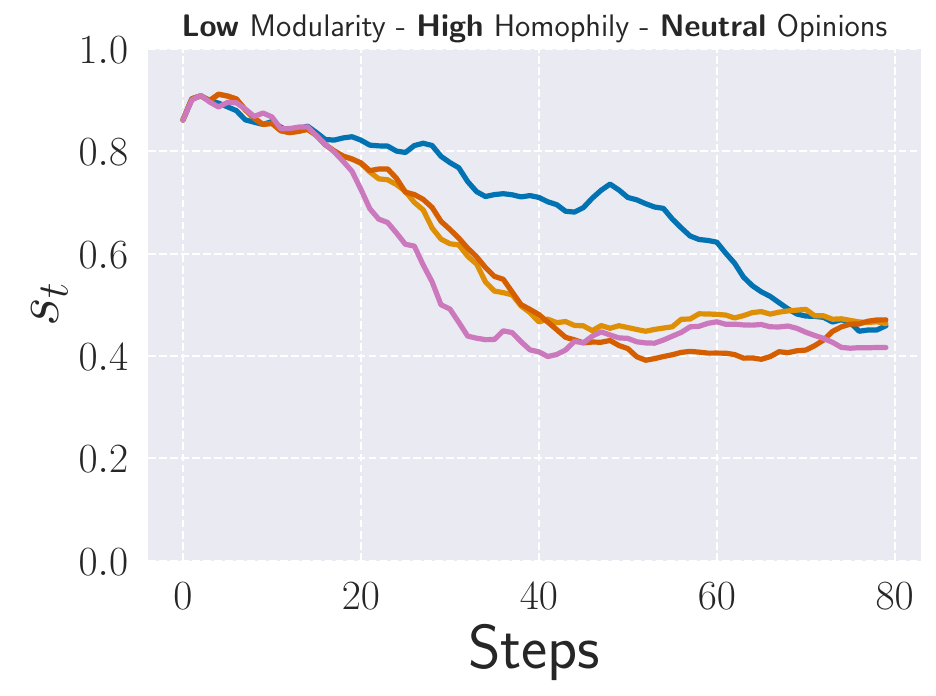} &
        \includegraphics[width=0.22\textwidth]{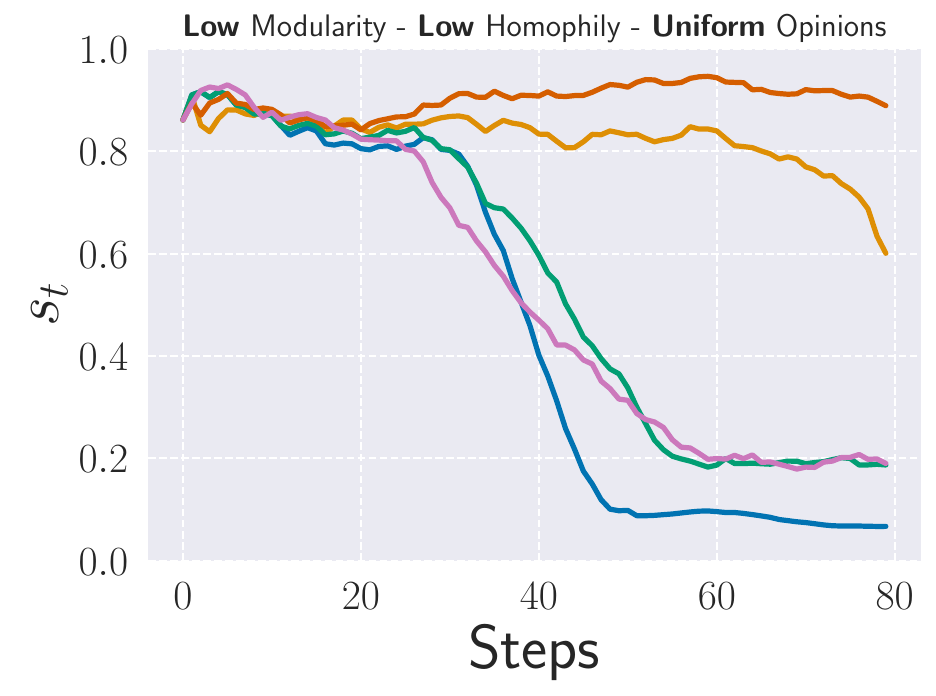} \\
    \end{tabular}
      \vspace{-4mm}
    \caption{Sentiment content $s_\content$ produced at each step of our fine-tuning procedure. Each column refers to a different opinion distribution: positive/negative/neutral/uniform (from left to right). Each plot depicts the trend varying the network structure in terms of modularity and homophily. Colors indicate different positions of the LLM agent.}
    \label{fig:synthetic_sentiment}
\vspace{-2mm}
\end{figure*}


\spara{Baselines.} We evaluate the effectiveness of our approach to fine-tuning the LLM, comparing it with the following baselines: BERT\footnote{https://huggingface.co/google-bert/}~\cite{kenton2019bert}, GPT-2~\cite{gpt2}, LLaMA3.1-70B-Instruct\footnote{\url{https://huggingface.co/meta-llama/Llama-3.1-70B-Instruct}}, and ChatGPT-4o\footnote{\url{https://openai.com/index/hello-gpt-4o/}}, in addition to the vanilla (not fine-tuned) Gemma-2B model~\cite{gemma}.
We invoke each model in inference to complete the same query $q$ used in the synthetic experiments (``Cats are the most'') and in the real-world scenarios (``[Brexit]/[2016 Italian constitutional referendum] is the most''). 
\added{When using BERT, we start from $q$ and masking the next word up to a final length of $10$. For all the other baselines, we set a maximum number of output tokens equal to $128$ and a temperature of $0.7$.}
Consistently with the fine-tuning process, we retrieve a number of responses equal to the batch size (8), and we further inject them into the network starting from the same source nodes. We average the results in terms of produced engagement. 

\noindent{\textbf{Reproducibility.}} Experiments were performed on an NVIDIA DGX with 4 V100(32Gb) GPUs. Code is publicly available\footnote{\url{https://github.com/mminici/Engagement-Driven-Content-Generation}}.


\section{Findings}
\label{sec:findings}
\subsubsection*{RQ 1: Can the LLM agent learn to generate content that maximizes engagement?\\}


The engagement $|\mathcal{A}|$ produced at each step of our fine-tuning procedure is depicted in Figure~\ref{fig:synthetic_engagement}, while the corresponding sentiment $\opinion_\content$ is reported in Figure~\ref{fig:synthetic_sentiment}.
Both figures exhibit a windowed moving average over 15 steps.
Each plot corresponds to a specific configuration defined by modularity, homophily, and network opinion, while colors denote different LLM positions, as described above\footnote{Not all the lines are reported whenever multiple configurations coincide.}.

In Figure~\ref{fig:synthetic_engagement}, the dashed lines indicate the maximum number of nodes that can be activated, as returned by $\mathcal{M}_{\epsilon}$ for the given configuration.
A comprehensive analysis of this engagement upper bound is provided in the Appendix, in Figure~\ref{fig:sent-network-response}.
The dotted lines represent engagement levels achieved when content is generated by the reference model, \(\llmm{\theta^{(0)}}\), as described in Algorithm~\ref{alg:procedure} (i.e., the LLM without fine-tuning). The complete set of experiments on the synthetic network is publicly accessible\footnote{\url{https://shorturl.at/2REXQ}}.

\spara{Finding 1: The LLM agent can maximize engagement in environments with positive opinions.}
First, we can observe that we reach a faster convergence in almost all the experiments when positive opinions are considered (left column). A slower trend occurs only when the LLM agent is placed in the smallest community, making the nodes harder to reach.

While this behavior on positive opinions might be expected (since open-source LLMs are biased towards generating user-friendly content), we observe that most configurations result in significantly higher engagement compared to the lower bound (which indeed represents the baseline LLM with no fine tuning). This improvement is particularly evident in settings with low modularity and when the LLM is placed in the \texttt{Echo-high} community. The sentiment of the generated content remains consistently positive, as shown in the corresponding plot in Figure~\ref{fig:synthetic_sentiment}.
\spara{Finding 2: The LLM agent can maximize engagement even in adversary opinion configuration.}
In adverse environments where users' opinions are skewed towards negative sentiments (second column in Figure~\ref{fig:synthetic_sentiment}), the proposed framework enables the LLM to deviate from its natural tendency to spread positive content and instead produce high engagement with negative sentiment content.
This effect persists even when the underlying social graph is neutrally distributed (third column) or uniformly distributed (last column). The latter configuration is particularly surprising, as it demonstrates that the LLM agent can maximize engagement even when nodes' opinions are not skewed. In other words, it manages to find the optimal content to generate for activating as many users as possible, balancing across high-variance opinions.

\spara{Finding 3: Generated content aligns with the optimal sentiment.}
The framework enables the LLM agent to optimize engagement by aligning the sentiment of the generated content with the optimal sentiment for the environment. Convergence curves, as shown in Figure~\ref{fig:synthetic_sentiment}, demonstrate that the final sentiment adapts to the specific configuration and placement of the LLM agent.

\spara{Finding 4: Convergence is sensitive to the position of the LLM agent.}
Generally, the fine-tuning process succeeds within the fixed number of steps in most experiments. However, the LLM agent's position affects the convergence rate. In particular, high centrality tends to have an advantage, achieving faster convergence in nearly all configurations.
Few configurations do not reach convergence within the fixed number of steps. An example is the ``Comm-smallest'' configuration in networks that are uniformly distributed, with low modularity and homophily. This likely happens as starting with positive sentiment in a uniformly distributed and weakly connected community makes convergence more difficult.


\subsubsection*{RQ 2: Can the LLM agent generate realistic and meaningful content that is comparable to actual propagated content?\\}

\begin{figure*}[ht!]
    \centering
    \begin{minipage}[t]{0.45\linewidth}
        \centering
        \includegraphics[width=.8\linewidth]{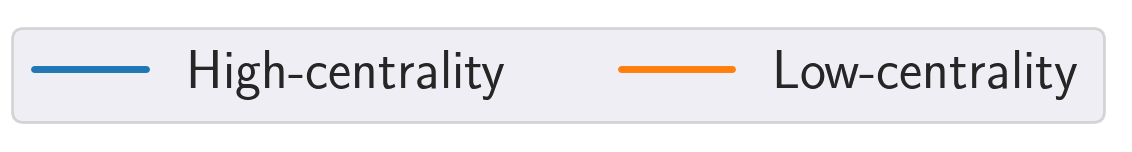}
    \end{minipage}
    \begin{tabular}{cccc}
    \centering
        \includegraphics[width=0.23\linewidth]{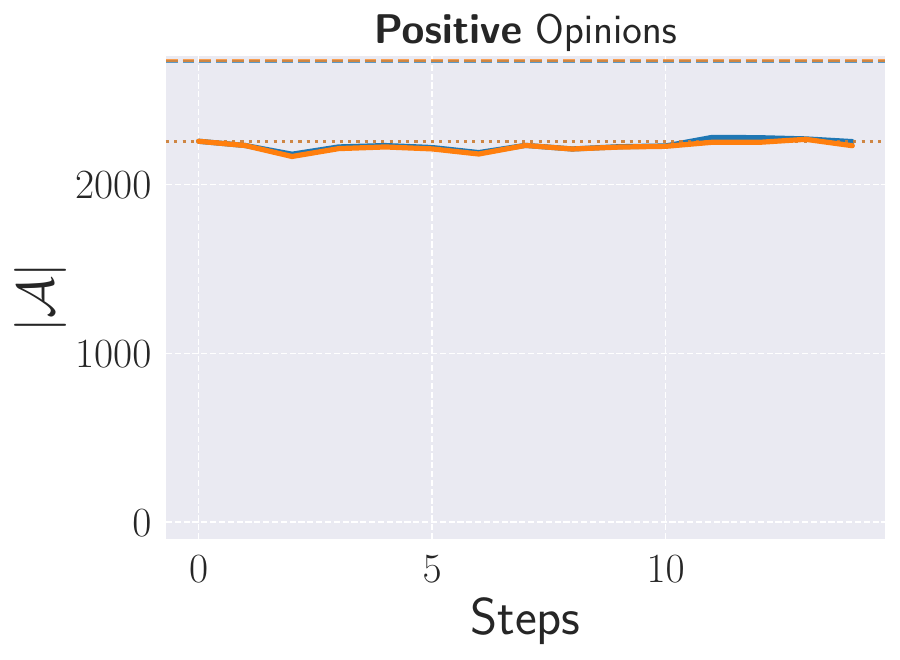} &
        \includegraphics[width=0.23\linewidth]{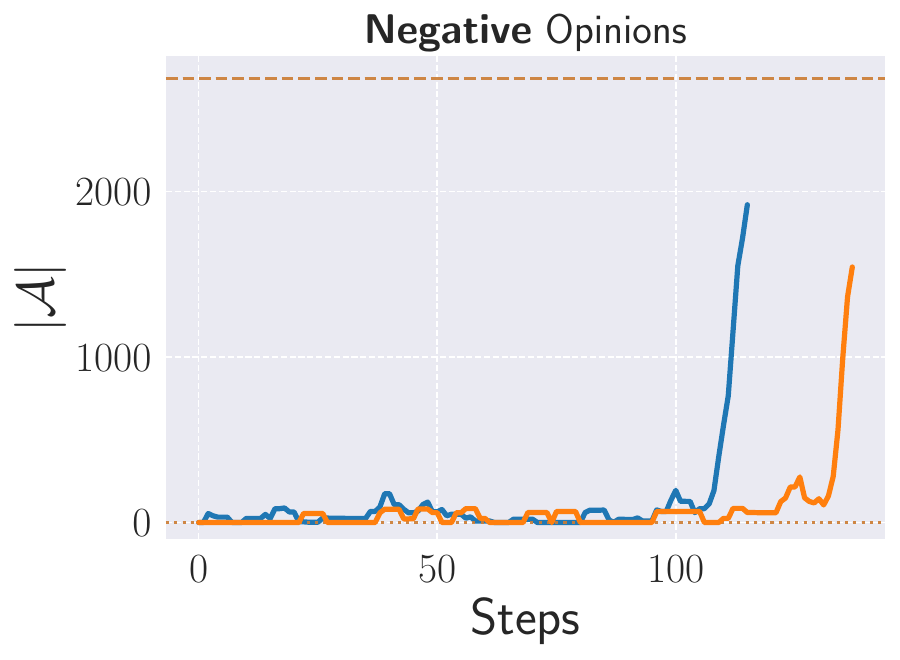} &
        \includegraphics[width=0.23\linewidth]{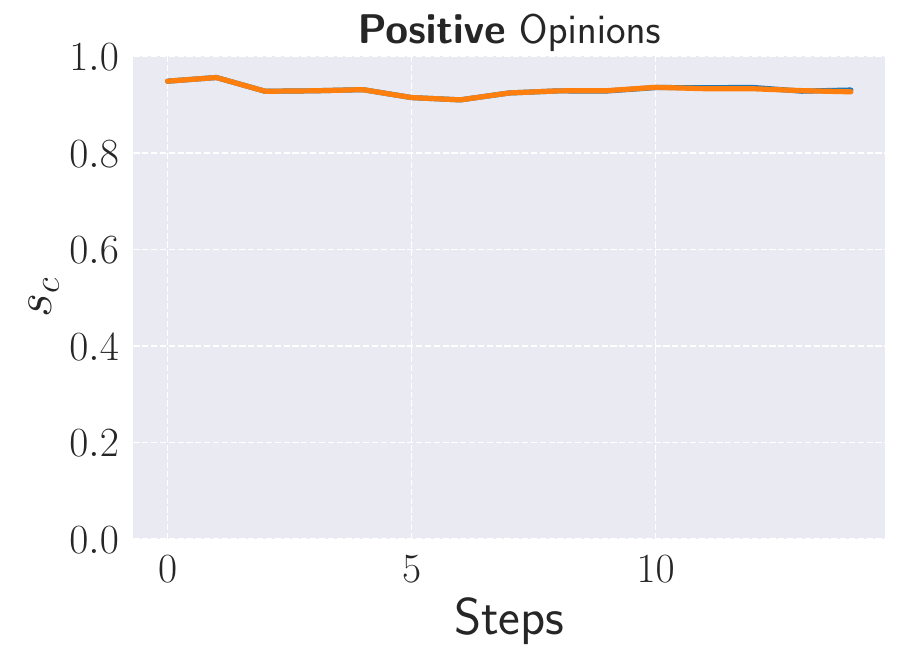} &
        \includegraphics[width=0.23\linewidth]{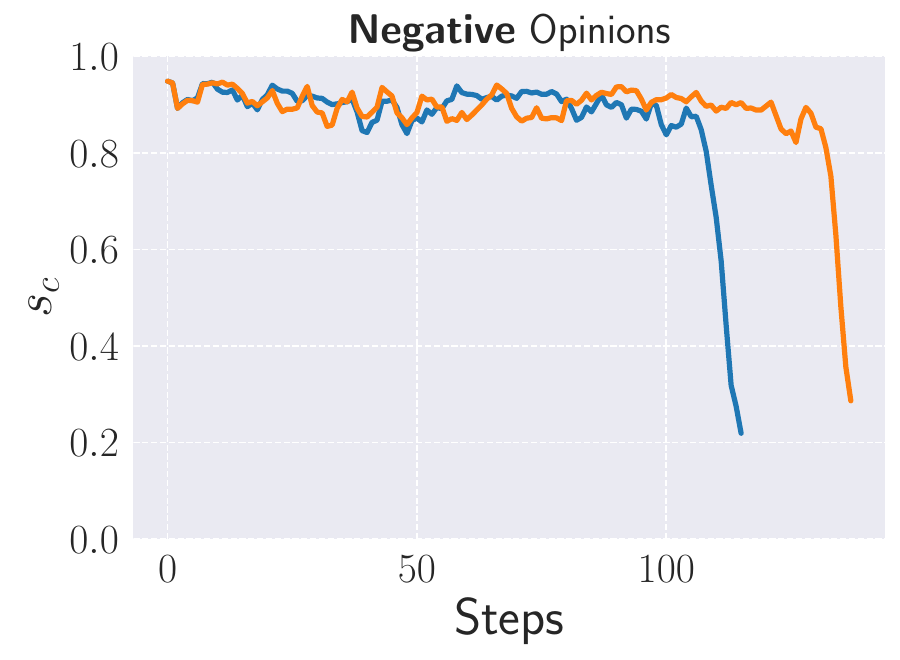} \\
    \end{tabular}
    \caption{Engagement (left) and Sentiment (right) produced at each step of our fine-tuning procedure over the \texttt{Brexit} communities (positively and negatively distributed). Colors represent different positions of the LLM agent, based on the betweenness centrality. Dashed lines represent the maximum engagement within that configuration.}
    \label{fig:brexit_results}
\end{figure*}

\begin{figure*}[ht!]
    \centering
    \begin{tabular}{cccc}
    \centering
        \includegraphics[width=0.23\linewidth]{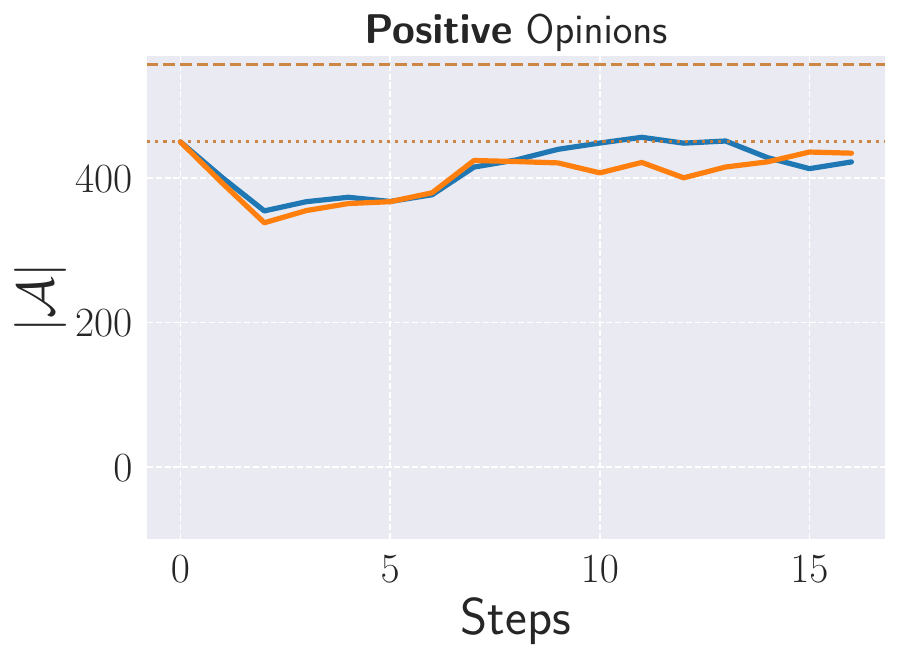} &
        \includegraphics[width=0.23\linewidth]{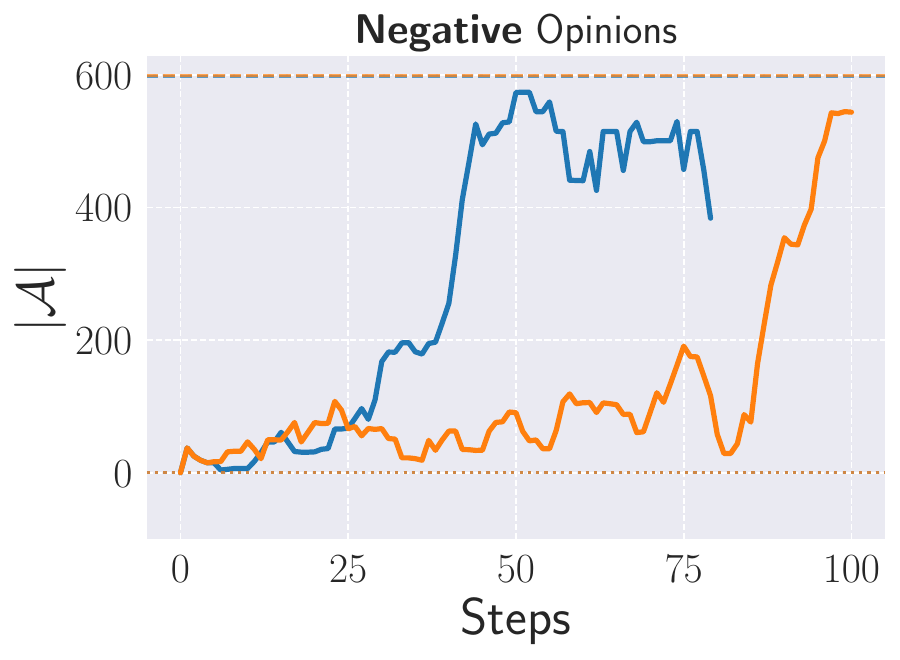} &
        \includegraphics[width=0.23\linewidth]{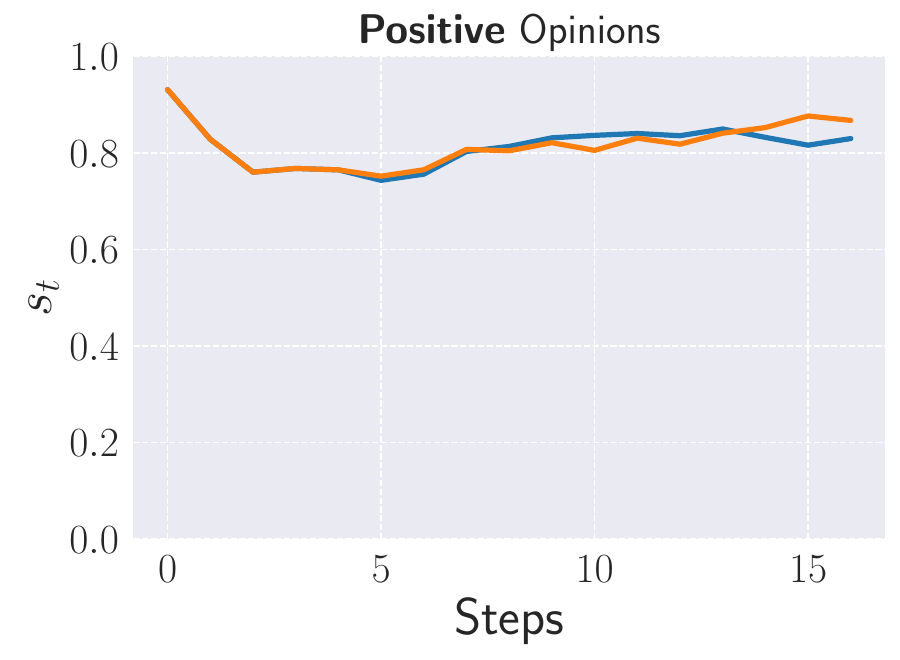} &
        \includegraphics[width=0.23\linewidth]{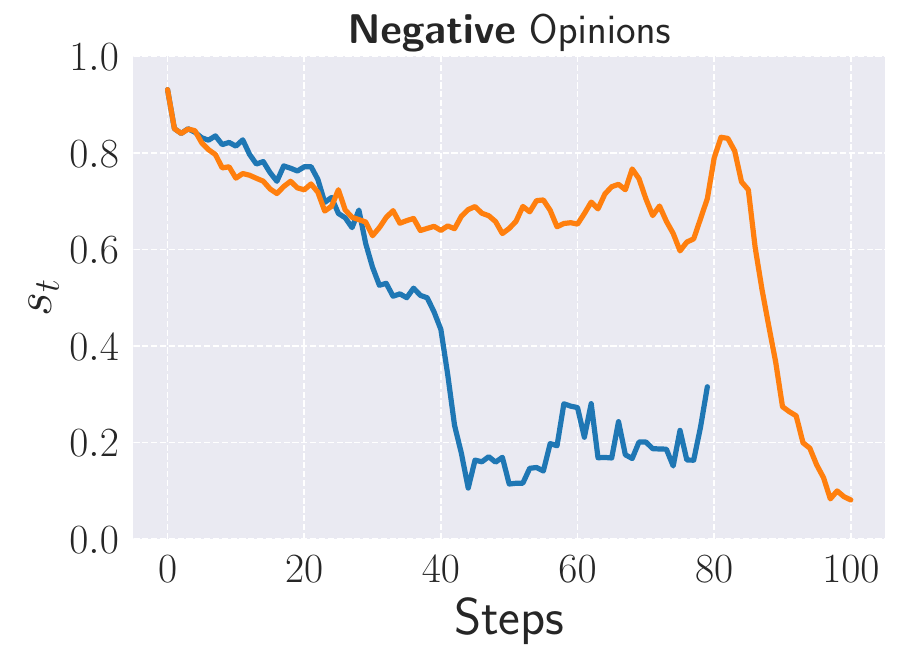} \\
    \end{tabular}
    \caption{Engagement (left) and Sentiment (right) produced at each step of our fine-tuning procedure over the \texttt{Italian Referendum} communities (positively and negatively distributed). Colors represent different position of the LLM agent, based on the betweenness centrality. Dashed lines represent the maximum engagement within that configuration.}
    \label{fig:referendum_results}

\end{figure*}

\spara{Finding 5: The LLM agent is effective in real-world environments.}
Preliminary experiments on the Brexit network revealed that the prominence of positive users biases content generation towards positive communities. 
To overcome this specific issue on real networks, we reformulated the goal of the experiment to determine if content generation can be tailored to engage specific subgroups of users representing polarized communities. We thus isolated the two largest communities of the Brexit network using the Louvain algorithm~\cite{louvain}: one positive (``Leave'', 3,095 nodes and 209,764 edges) and one negative (``Remain'', 2,894 nodes and 121,325 edges). We similarly processed the Italian Referendum network, thus obtaining a positive partition (``Yes'', 633 nodes and 20,528 edges) and a negative one (``No'', 616 nodes and 10,461 edges).

We then applied the learning procedure to each subgraph of each real network, positioning the LLM agent at the node with the highest or lowest betweenness centrality.


Figures~\ref{fig:brexit_results} and~\ref{fig:referendum_results} report the results of these experiments.
Consistently with the synthetic experiments, we observe rapid convergence when the network's opinion is predominantly positive and a slower convergence rate within the negative community. In both scenarios, the LLM adjusts the generated content to maximize engagement, eventually nearing the optimal threshold. This shows that our framework is effective, not only in a synthetic environment, but also in real-world contexts.
The execution time remains feasible, taking $\sim$$2$ minutes per optimization step for the real networks.


\begin{table}[t!]
    \centering
    \caption{Real vs. generated content on the selected communities. Each content is associated with the engagement generated over the whole population on \textsl{Brexit}.    \label{tab:tweet_comparison}}
    \vspace{-3mm}
    
    \resizebox{\columnwidth}{!}{
    \begin{tabular}{cp{5cm}p{1cm}p{5cm}p{1cm}}  
        \toprule
         {} & \centering{\textbf{\textcolor{red}{Leave}}} & $|\mathcal{A}|$ & \centering{\textbf{\textcolor{blue}{Remain}}} & $|\mathcal{A}|$  \\
        \midrule
         \rotatebox[origin = c]{90}{\hspace{-1.6cm} \textbf{Real}} & \texttt{Truly the most damning fact about about \#EuRef is that our democracy is broken.  Only \#VoteLeave will fix it.}
            & 2,700 (35,6\%)  &
                    \texttt{We are Great Britain not little England, let’s keep our place in the world and not head into recession \#VoteRemain \#StrongerIn \#ITVEURef}
            &  3,739 (49,3\%)\\
        \midrule
        \rotatebox[origin = c]{90}{\hspace{-6cm} \textbf{Generated}} &
            \texttt{Brexit is the most significant turning point in the UK's history. It's not just about trade any more, but about the entire social fabric of the nation. Join us on this historical journey as we explore the key moments and significant events leading up to Brexit.}
        & 2,799 (36,9\%) &
            \texttt{Brexit is the most gravity issue facing the UK.
            Key arguments against Brexit:
            \begin{itemize}[leftmargin=*]
                \item Trade barriers between the UK and the EU would cause significant disruption and damage to the economy.
            \item The UK would no longer have access to the single market, which would limit its access to free movement of goods, services and capital.
            \item Brexit would be bad for jobs, with the loss of thousands of jobs in key industries such as manufacturing.
             \end{itemize}}
         & 2,422 (31,9\%) \\
        \bottomrule
    \end{tabular}
    }

\vspace{3mm}

    \caption{Real vs. generated content on the selected communities. Each content is associated with the engagement generated over the whole population on \textsl{Italian Referendum}.    \label{tab:tweet_comparison_referendum}}
       \vspace{-3mm}
    \resizebox{\columnwidth}{!}{
    \begin{tabular}{cp{5cm}p{1cm}p{5cm}p{1cm}}  
        \toprule
         {} & \centering{\textbf{\textcolor{red}{No}}} & $|\mathcal{A}|$ & \centering{\textbf{\textcolor{blue}{Yes}}} & $|\mathcal{A}|$  \\
        \midrule
         \rotatebox[origin = c]{90}{\hspace{-1.2cm} \textbf{Real}} & \texttt{Don't trust the polls, never ease up on the fight until the end!
Let’s vote and encourage others to vote \#NO! \#IVoteNO}
            & 2,233 (77,2\%) &
                    \texttt{If No wins, nothing changes. With Yes, we take a step forward to restart Italy \#IVoteYes}
            & 435 (15\%) \\

        \midrule
        \rotatebox[origin = c]{90}{\hspace{-7cm} \textbf{Generated}} &
            \texttt{2016 Italian constitutional referendum is the most concerning because:
            \begin{itemize}[leftmargin=*]
                \item It is contradicted by the Italian Constitution.
                \item It violates the principles of popular sovereignty and constitutionality.
                \item The outcome can not be predicted with certainty by the people.
            \end{itemize}.
            }
        & 2,233 (77,2\%) &
        \texttt{2016 Italian constitutional referendum is the most relevant example of a democratic constitutional due to the following aspects:
        \begin{itemize}[leftmargin=*]
        \item The constitutional referendum is a direct involvement of citizens in defining the rules of their communities and exercising their rights through direct participation.
        \item It is the first step towards the creation of a new legal order and shaping the relationship between public authorities and individuals.
        \item It allows the community to express its will.
        \end{itemize}
        }
         & 412 (14,2\%)

         \\
        \bottomrule
    \end{tabular}
    }
\end{table}

\spara{Finding 6: Generated content is fluent and polarized as the real Tweets and exhibits comparable engagement levels.}
Tables~\ref{tab:tweet_comparison} and~\ref{tab:tweet_comparison_referendum} show some examples of generated content, compared to real tweets that were shared on the Brexit and Italian Referendum networks, respectively.
We can see that the sentiment of the generated content is aligned with that of the real one. The notable difference is the length of the generated content, which is not controlled in our framework and departs from the typical length of real tweets, ranging within 280 characters.

The final set of experiments we perform is aimed at measuring how realistic is the potential engagement of the generated content.
As a first step, we assess whether the engagement model proposed in Section~\ref{sec:engagement-model} represents an adequate proxy for actual engagement on \textsl{Brexit}. We can in fact, witness (Figure~\ref{fig:correlation-engagement-brexit} in the Appendix) a positive correlation between the tweets with the highest propagation rate within the network, and the corresponding predicted engagement.


Next, we compare the predicted engagement from both real and generated content in Tables~\ref{tab:tweet_comparison} and~\ref{tab:tweet_comparison_referendum}.
To do this, we propagated each content across the entire network from the same starting position—the actual node that posted the real tweet. This approach ensures a fair comparison under identical initial conditions. The results show that the engagement levels for generated content are highly comparable to those for human text.

\begin{figure}
    \centering
    \begin{tabular}{cc}
    \includegraphics[width=0.48\columnwidth]{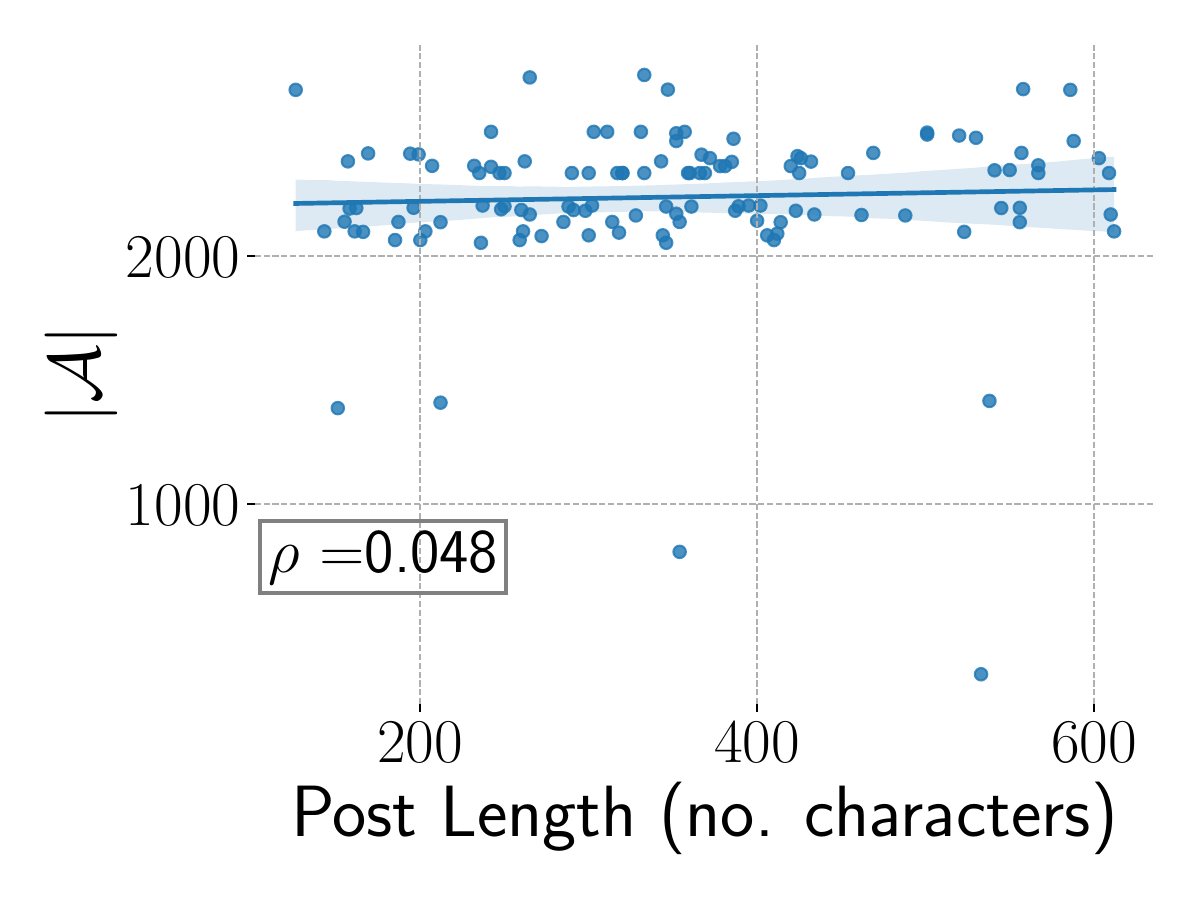} &
    \includegraphics[width=0.48\columnwidth]{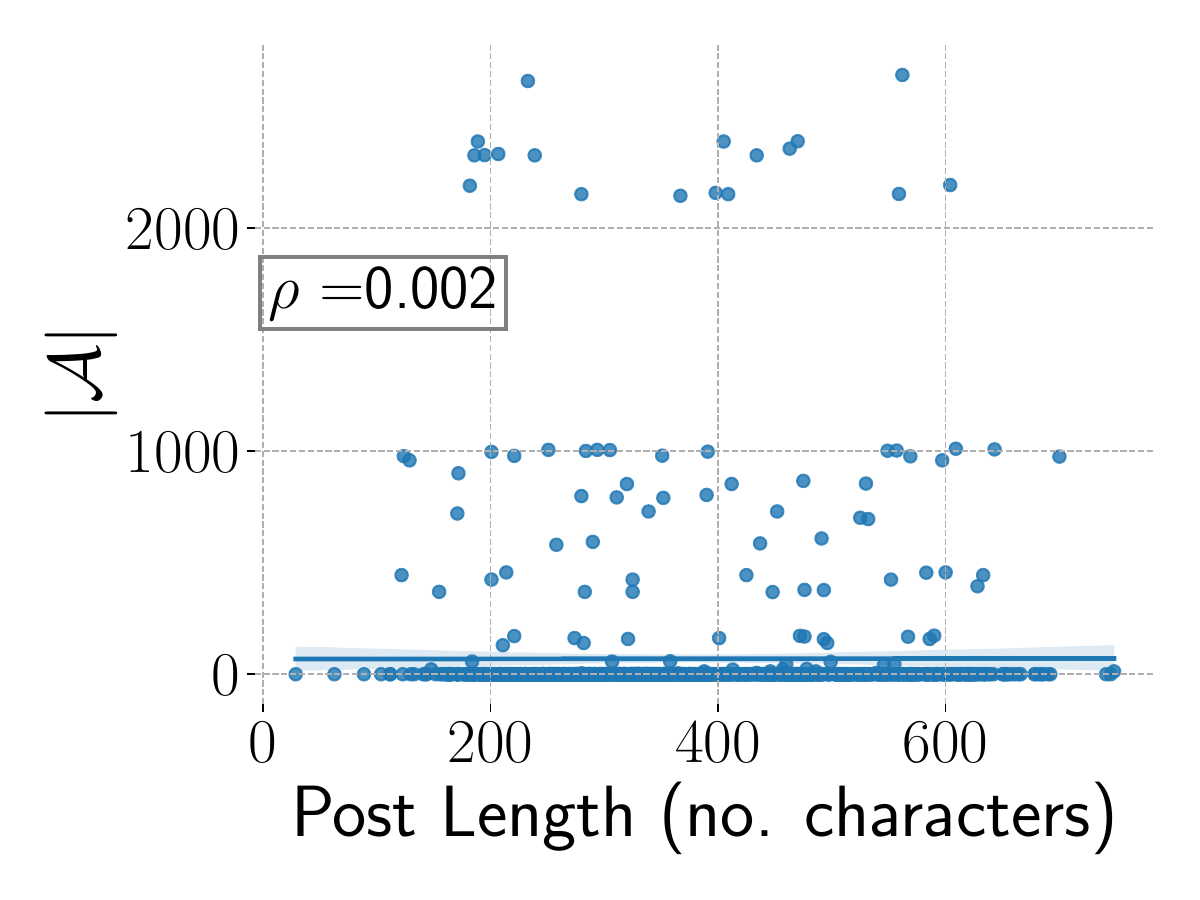}
    \end{tabular}
    \caption{Correlation between the produced engagement $|\mathcal{A}|$ and the post length (in terms of number of characters). The results are computed over the real Brexit network with positive opinions (left) and negative opinions (right).\label{fig:prompt_length-engagement-correlation}}
\end{figure}

\spara{Finding 7: The produced engagement does not depend on content length.} We further examined whether a correlation exists between the length of the generated post and the engagement it produces across the network. This analysis stems from the intuition that longer content might increase the likelihood of user engagement. However, as illustrated in Figure~\ref{fig:prompt_length-engagement-correlation}, our analysis on the Brexit dataset suggests no correlation between post length (measured in characters) and the number of users who engaged with it.

\begin{figure*}[ht!]
    \centering
    \begin{minipage}[t]{\linewidth}
    \centering
       \includegraphics[width=0.8\linewidth]{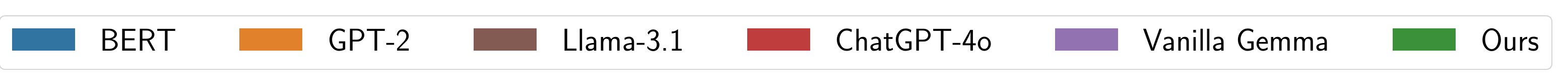}
    \end{minipage}
    \vspace{-0.3cm}
    \addtolength{\tabcolsep}{-0.8em}
    \begin{tabular}{cccc}
        \includegraphics[width=0.25\textwidth]{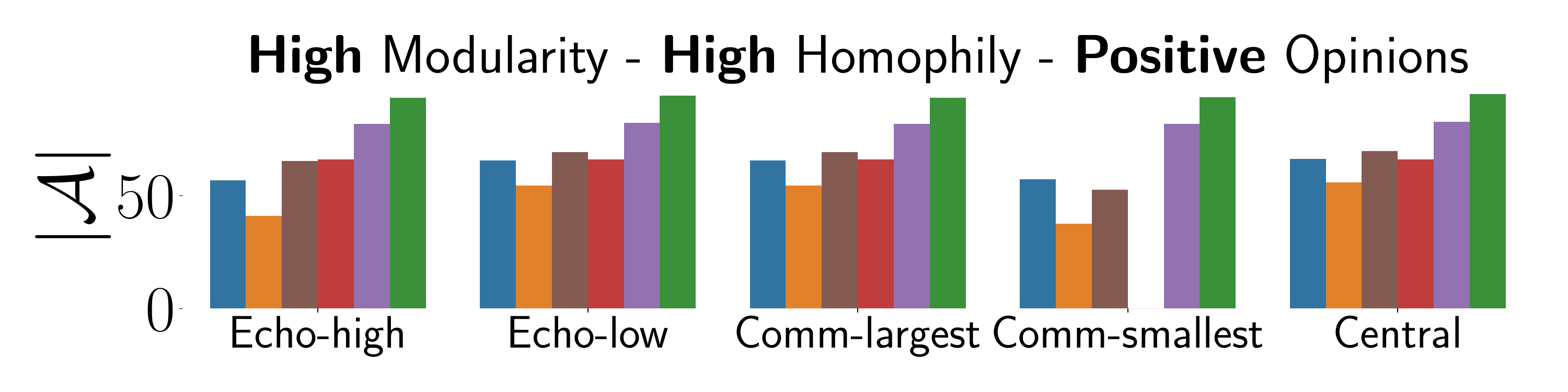} &
        \includegraphics[width=0.25\textwidth]{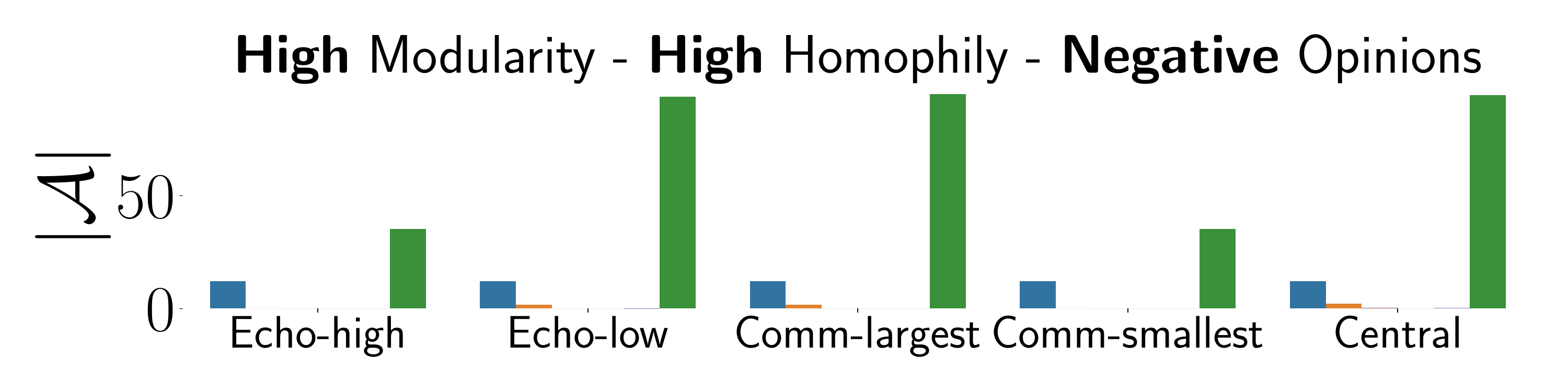} &
        
        \includegraphics[width=0.25\textwidth]{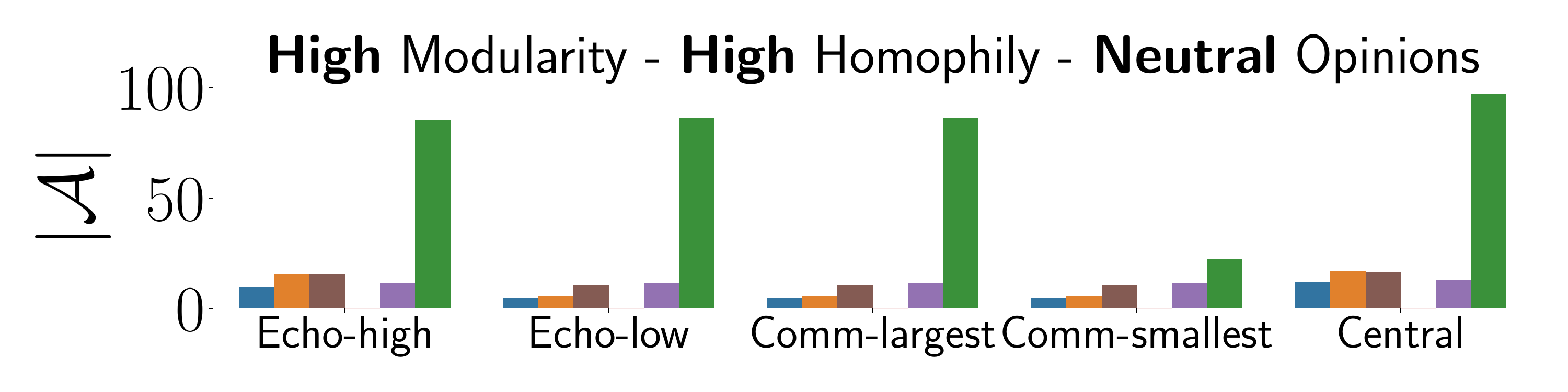} &
        \includegraphics[width=0.25\textwidth]{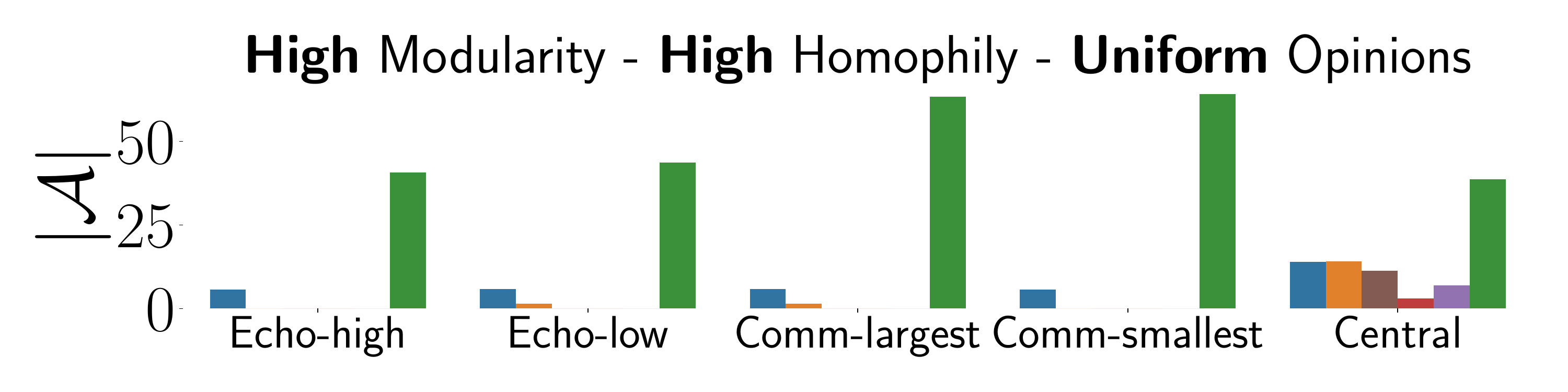} \\
        \includegraphics[width=0.25\textwidth]{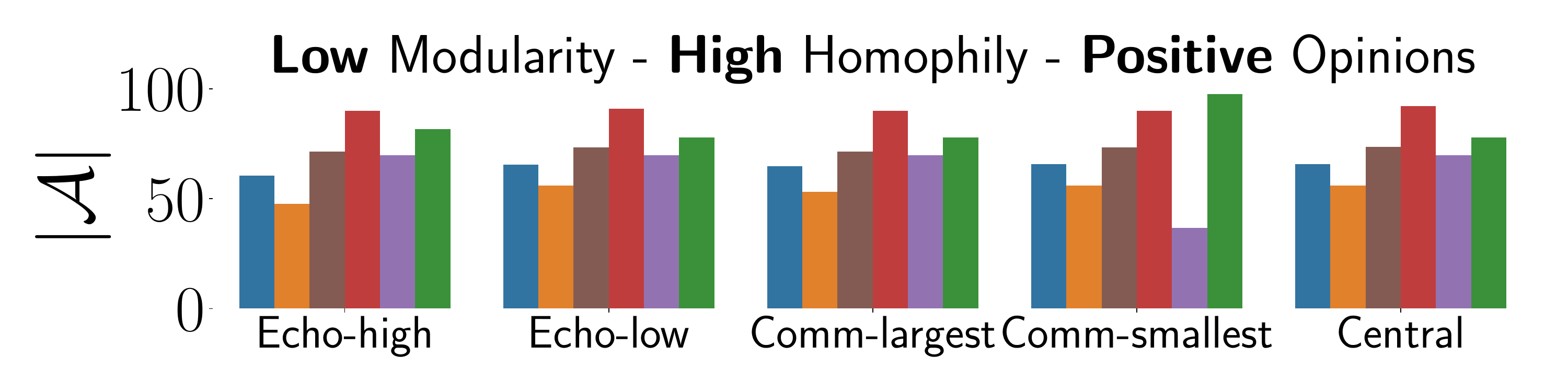} &
        \includegraphics[width=0.25\textwidth]{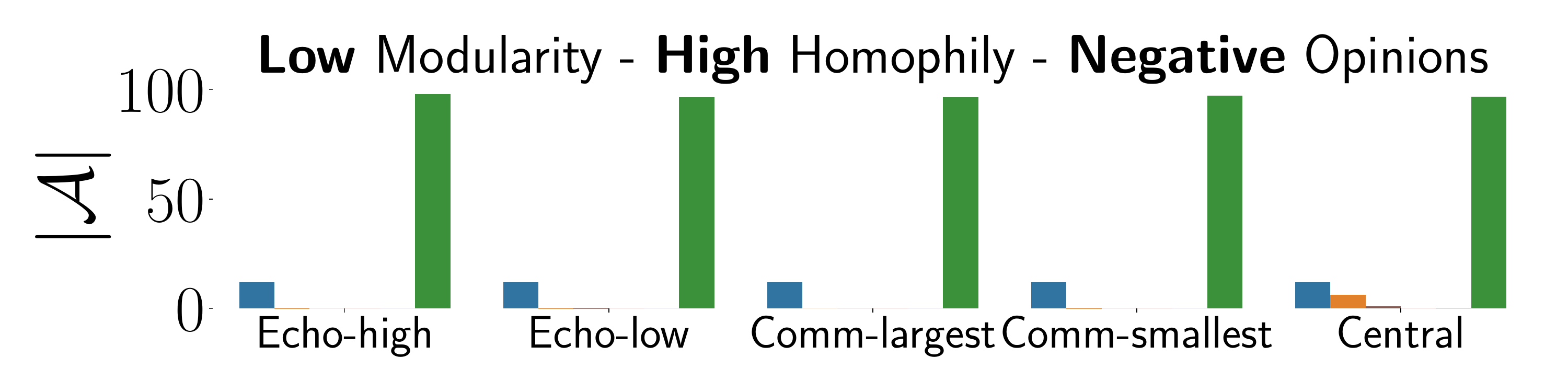} &
        \includegraphics[width=0.25\textwidth]{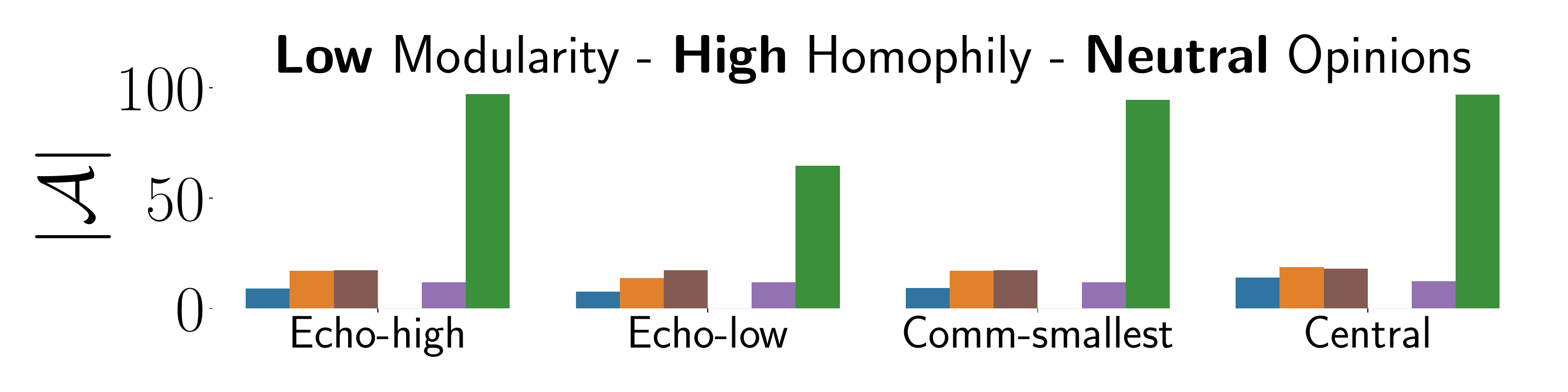} &
        \includegraphics[width=0.25\textwidth]{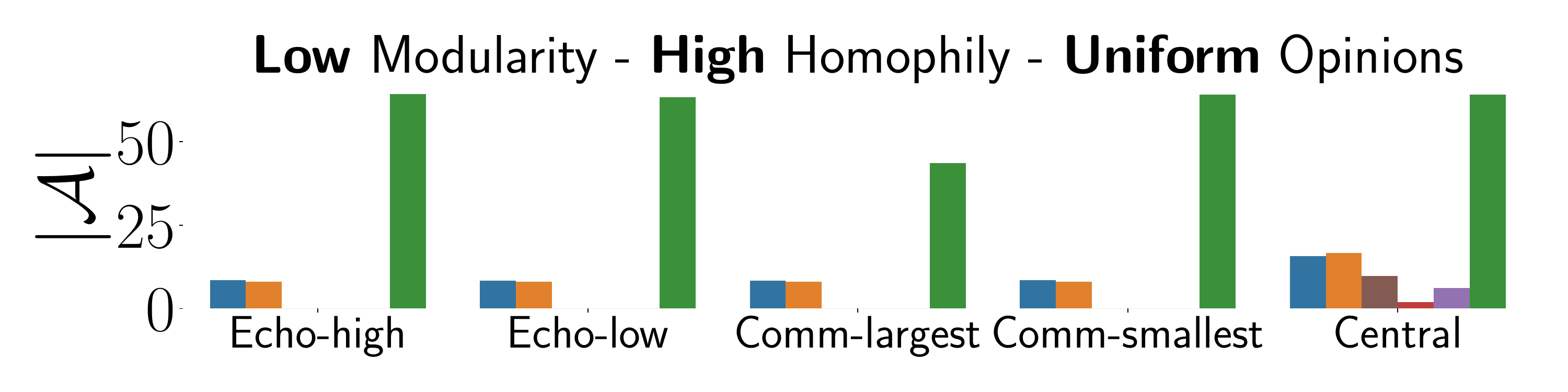} \\
        
        \includegraphics[width=0.25\textwidth]{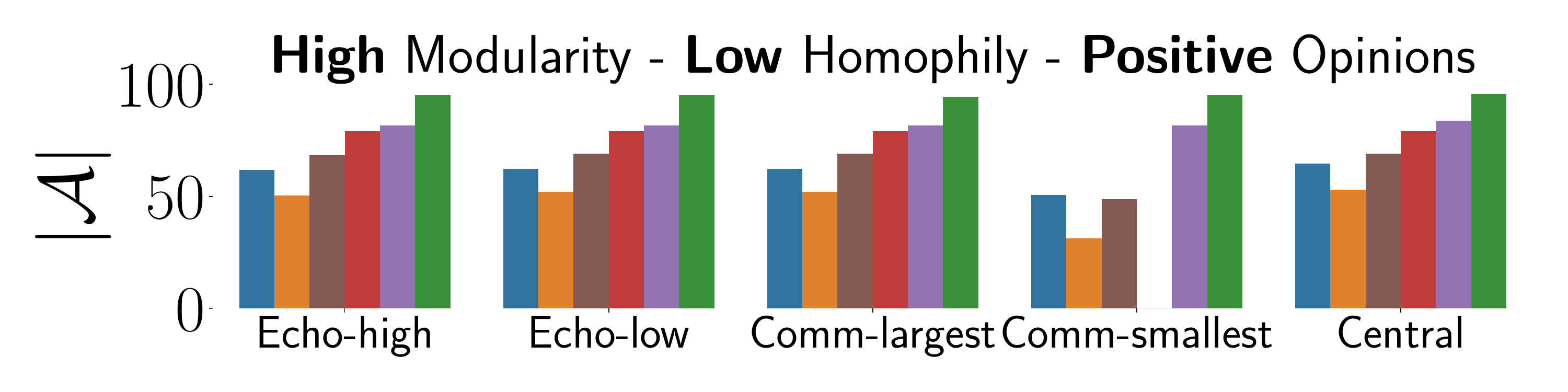} &
        \includegraphics[width=0.25\textwidth]{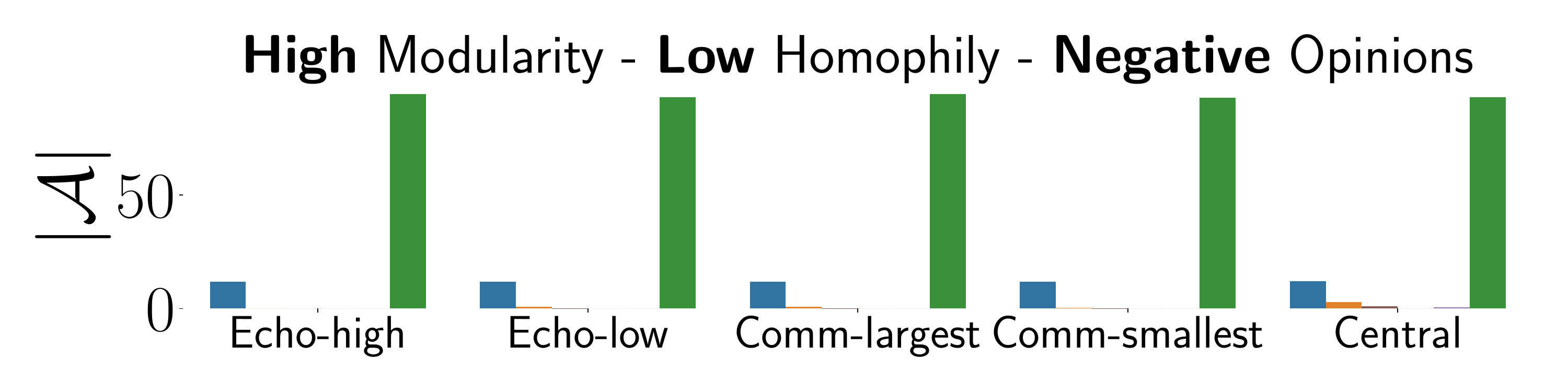} &
        \includegraphics[width=0.25\textwidth]{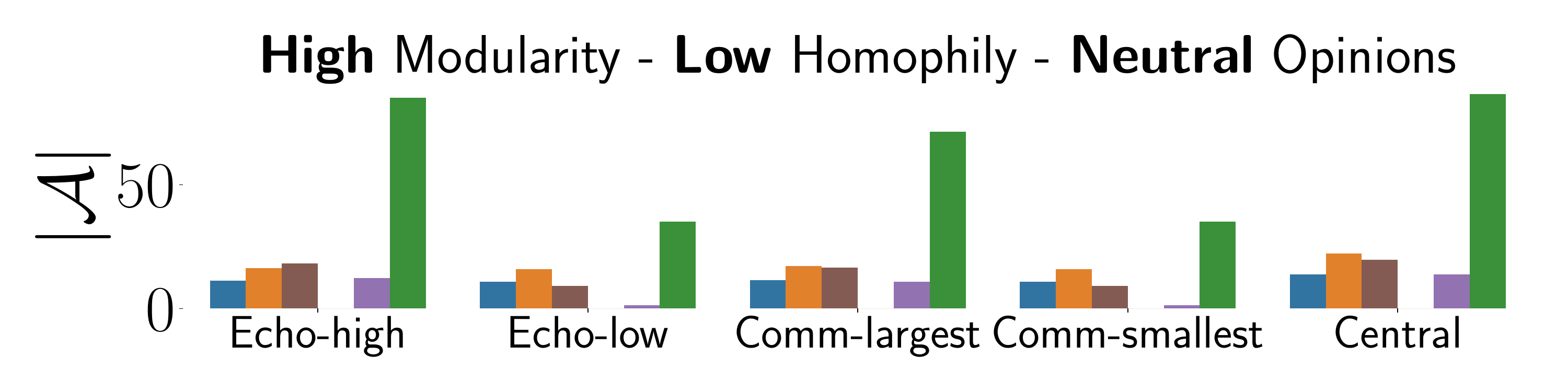} &
        \includegraphics[width=0.25\textwidth]{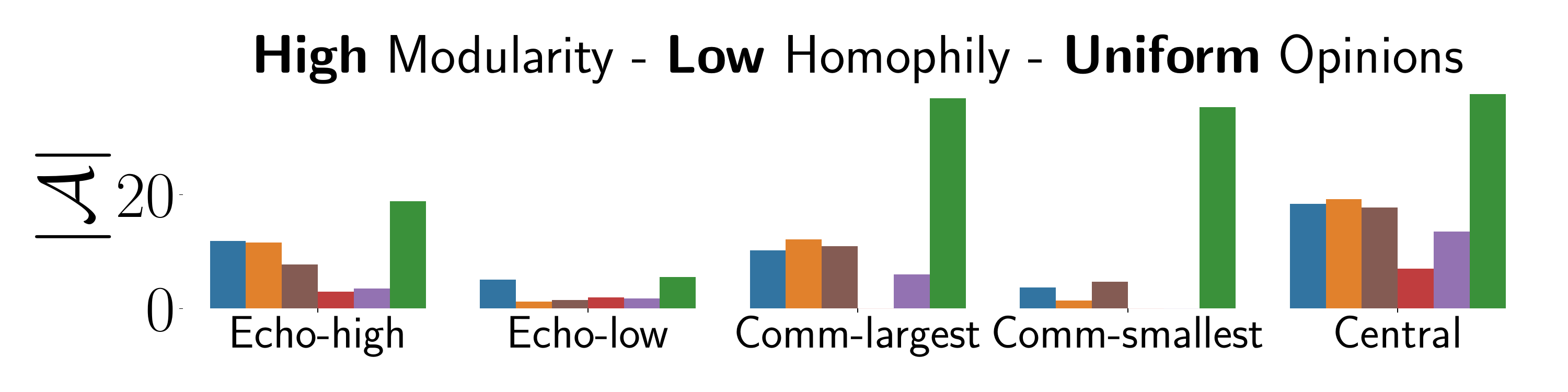} \\
        
        \includegraphics[width=0.25\textwidth]{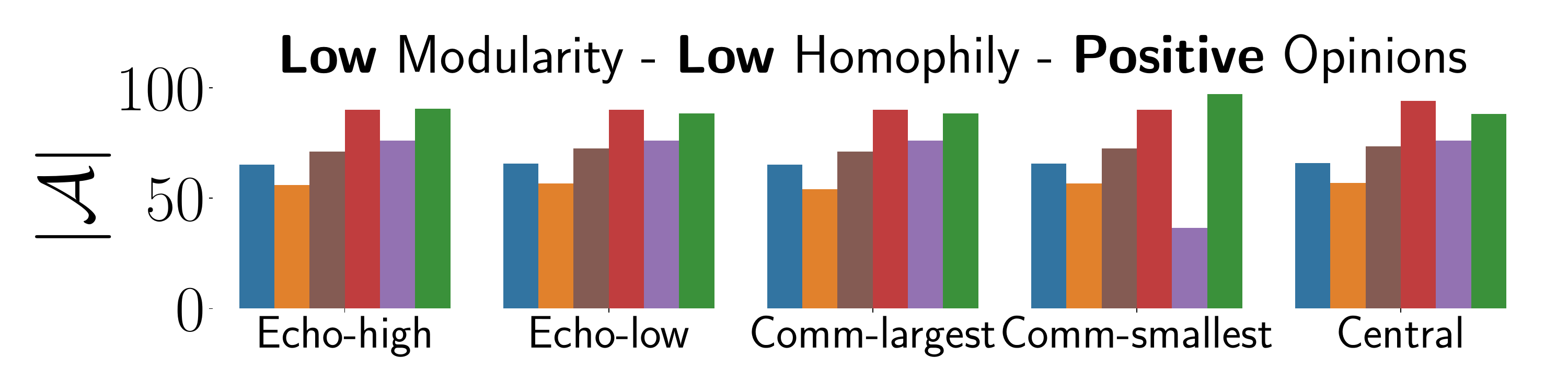} &
        \includegraphics[width=0.25\textwidth]{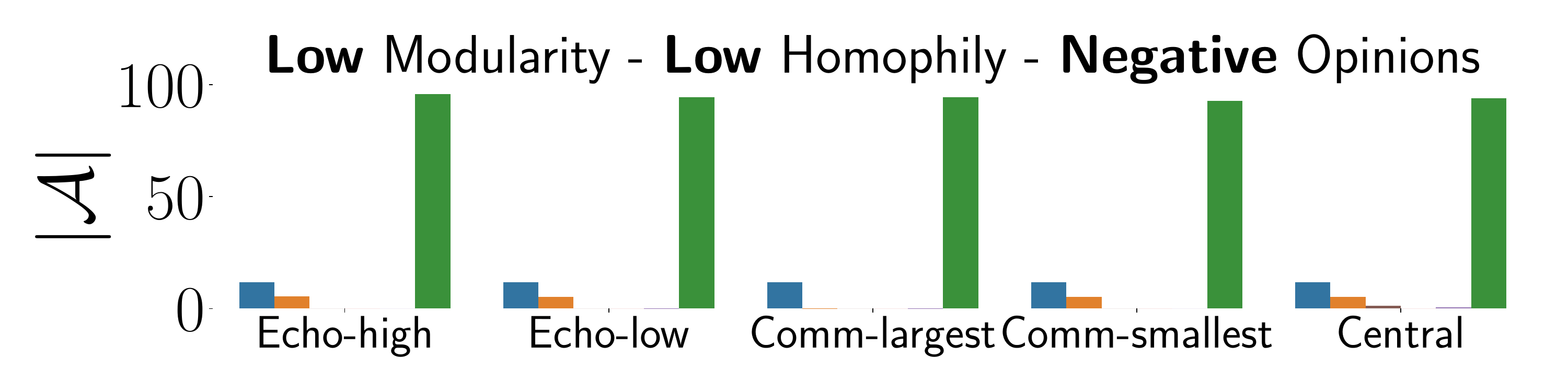} &
        \includegraphics[width=0.25\textwidth]{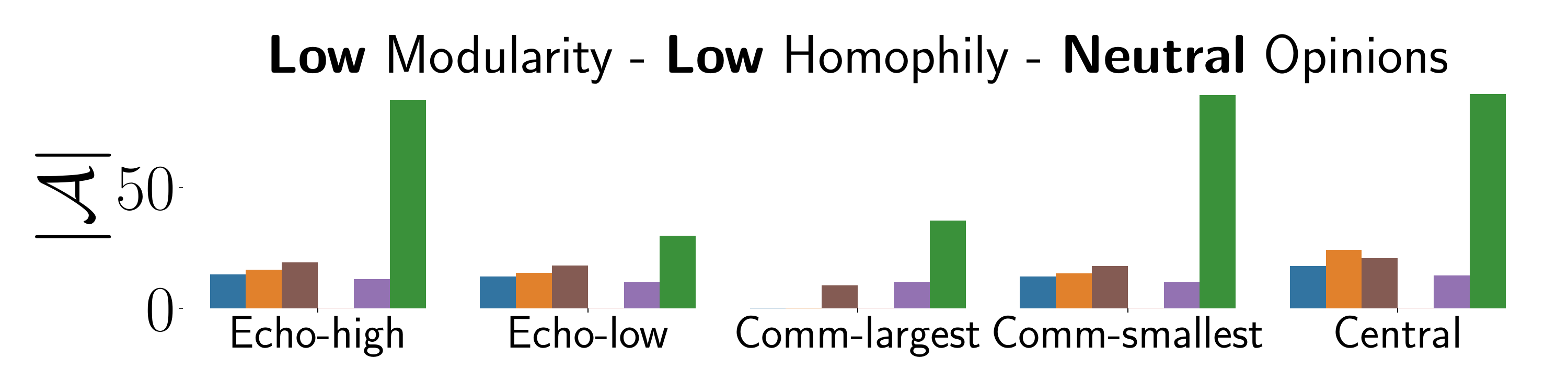} &
        \includegraphics[width=0.25\textwidth]{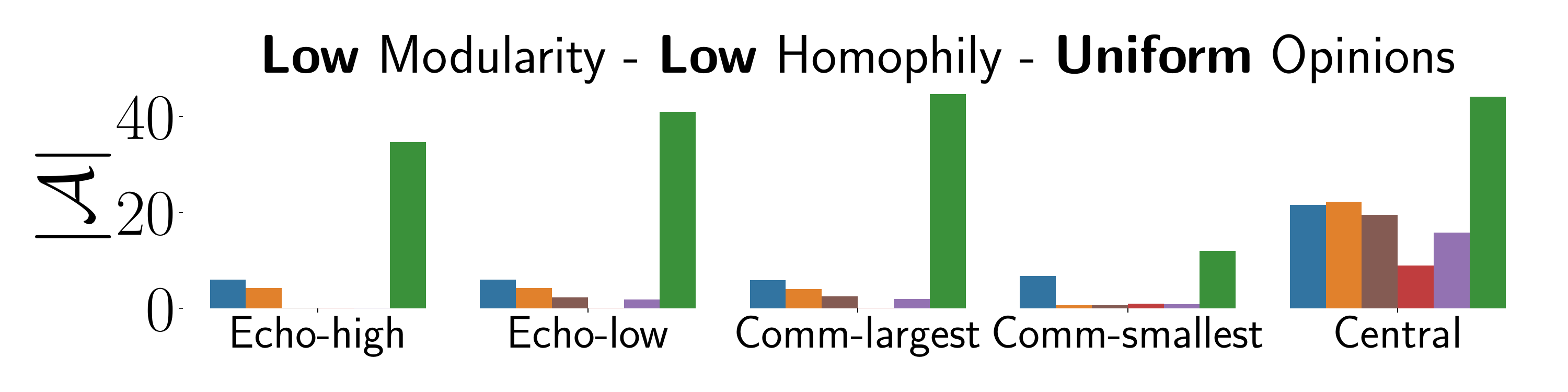} \\
    \end{tabular}
    \caption{Engagement $|\mathcal{A}|$ produced on the synthetic network by several baselines beyond the fine-tuned model. Each bar group represents a different position of the LLM agent within the social graph.}
    \label{fig:synthetic_engagement_baselines}
\end{figure*} 


\begin{figure}[ht!]
    \centering
    \begin{tabular}{cc}
        \includegraphics[width=0.425\columnwidth]{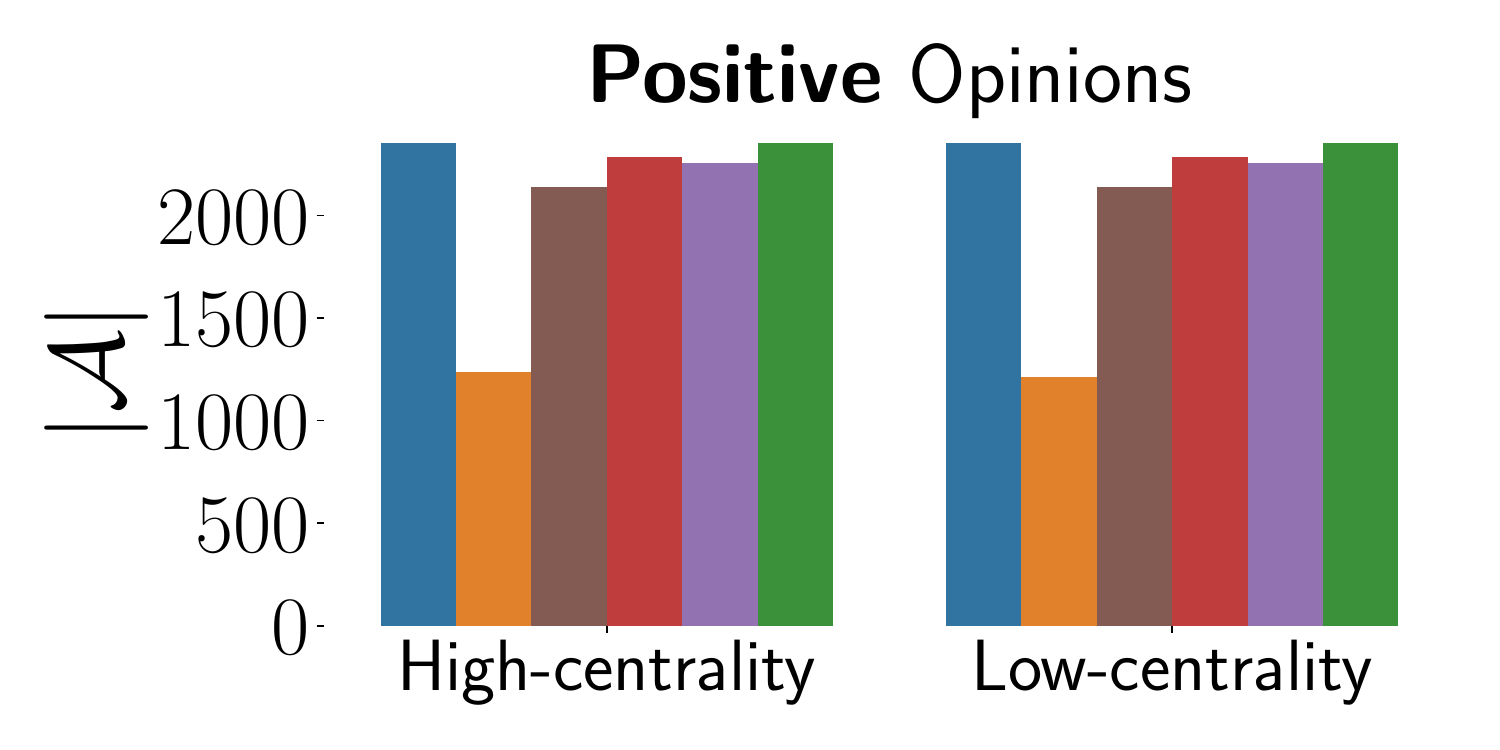} &
        \includegraphics[width=0.425\columnwidth]{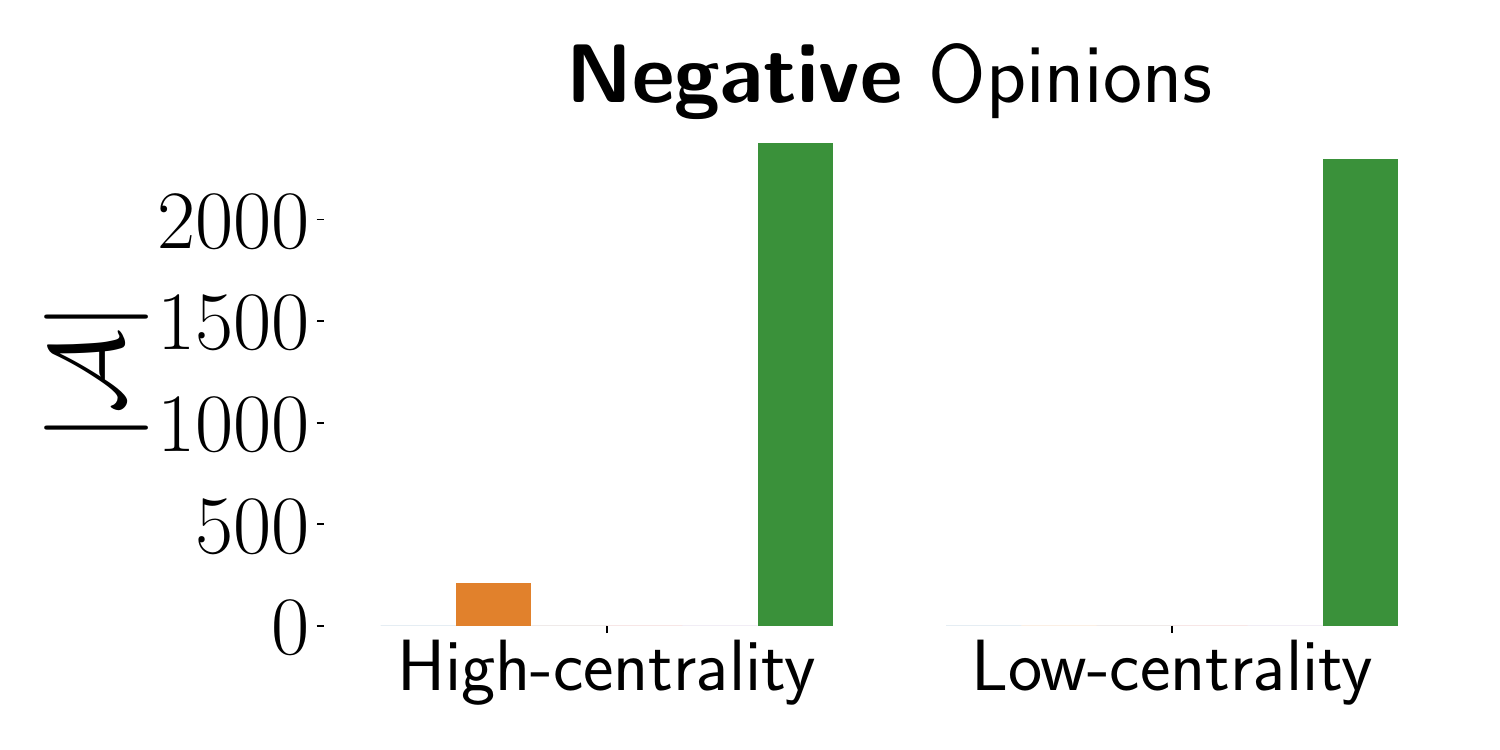} \\

    \end{tabular}
    \vspace{-3mm}
    \caption{Engagement $|\mathcal{A}|$ produced on the real Brexit network by several baselines beyond the fine-tuned model. Each bar group represents a different position of the LLM agent within the social graph.}
    \label{fig:brexit_engagement_baselines}
    \vspace{-3mm}
\end{figure}

\subsubsection*{RQ3: How does the proposed framework compare to other content generation methods from the literature?\\}
\begin{figure}[t!]
    \centering
    \begin{tabular}{cc}
        \includegraphics[width=0.425\columnwidth]{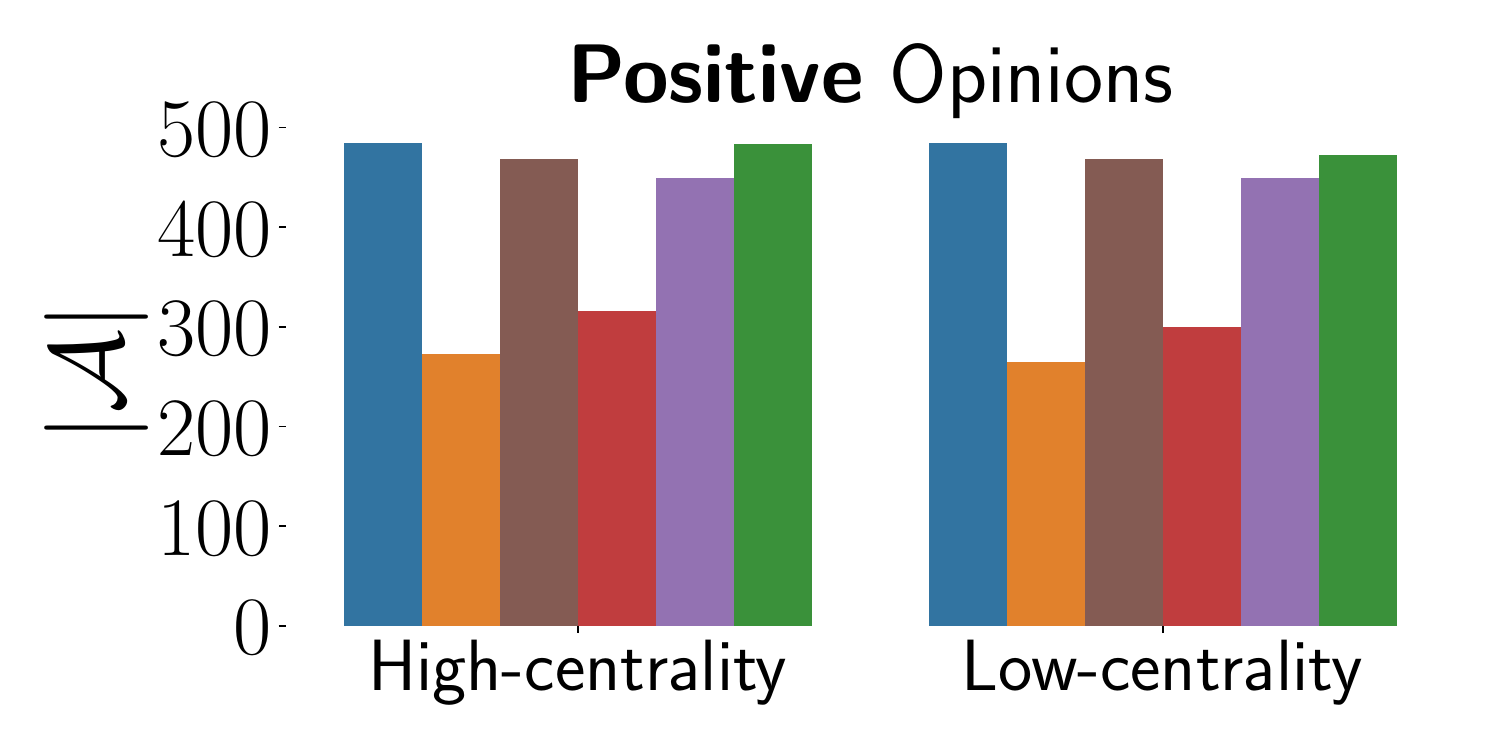} &
        \includegraphics[width=0.425\columnwidth]{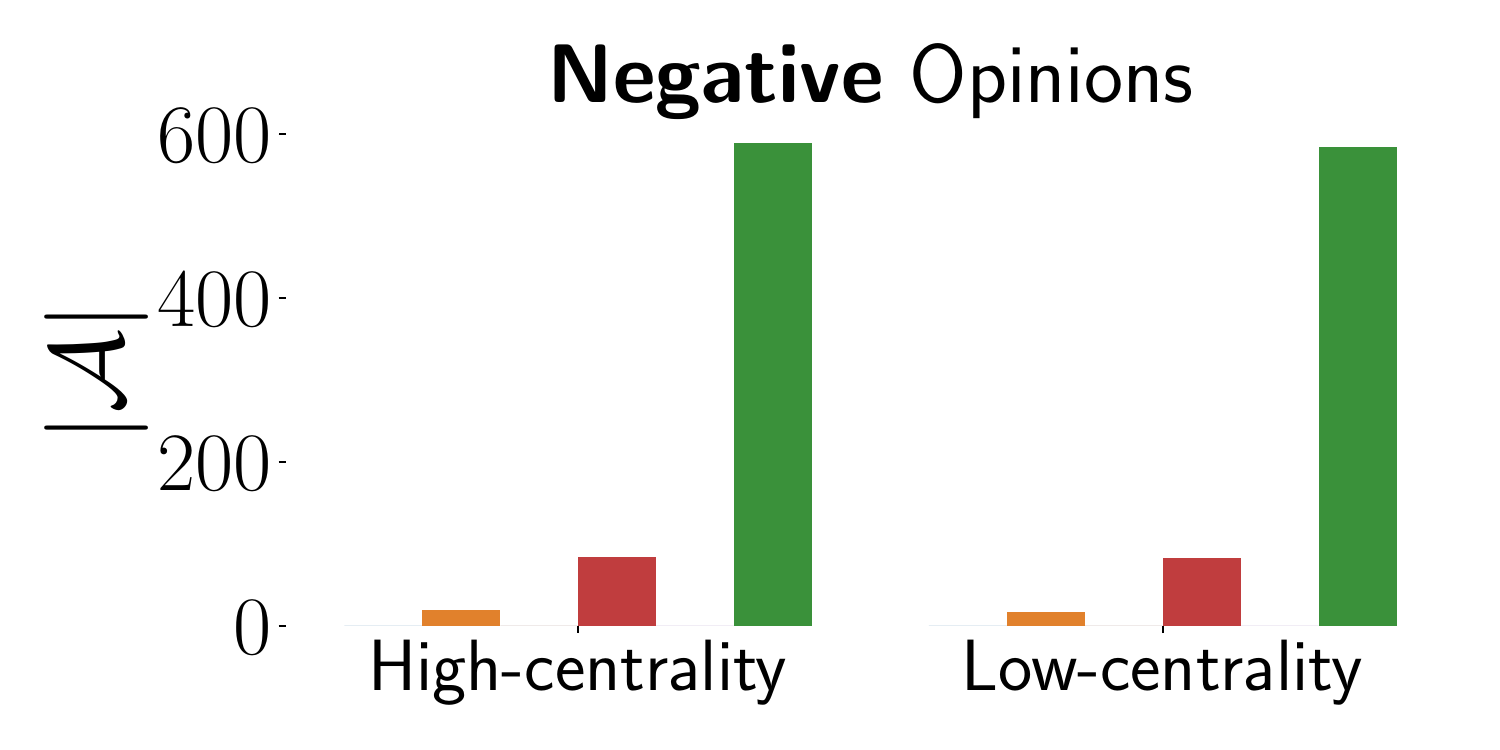} \\

    \end{tabular}
    \vspace{-3mm}
    \caption{Engagement $|\mathcal{A}|$ produced on the real Italian Referendum network by several baselines beyond the fine-tuned model. Each bar group represents a different position of the LLM agent within the social graph.}
    \label{fig:referendum_engagement_baselines}
    \vspace{-3mm}
\end{figure}

\spara{Finding 8: Our approach overcomes baseline and advanced language models.}
Using the same prompts as in the synthetic and real-world experiments, we compared the engagement generated by the baseline models previously described to that achieved by our fine-tuned LLM agent. Figure~\ref{fig:synthetic_engagement_baselines} illustrates the results on synthetic data across the considered configurations, while Figures~\ref{fig:brexit_engagement_baselines} and~\ref{fig:referendum_engagement_baselines} show the engagement obtained for the Brexit and Italian Referendum networks, respectively. 

As shown in the figures, while the fine-tuned LLM performs comparably to other models in scenarios where the opinion distribution is positively skewed (both synthetic and real-world), it significantly outperforms all baselines in cases where the network opinion is negative, neutral, or uniformly distributed. Notably, this improvement is evident not only over traditional language models like BERT and legacy LLMs such as GPT-2 but also over state-of-the-art, significantly larger models like LLaMA3.1-70B and ChatGPT-4o.

These results underscore two key findings: (i) our methodology is highly adaptable to varying network configurations, and (ii) fine-tuning the agent is essential to maximize engagement, particularly when user opinions are not mainly positive toward the topic.

\section{Discussion and Conclusions}
\label{sec:conc}
The goal of this paper is to present a framework that enables any LLM to generate content optimized for effective diffusion on social networks. We achieve this by designing a fine-tuning strategy that incorporates simulated feedback from an engagement model while preserving the natural fluency of the language. Through extensive experiments on both synthetic and real datasets, we demonstrate that the optimized LLM (after applying our Alg. \ref{alg:procedure}) successfully achieves this goal, producing viral yet realistic content under a wide range of possible initial conditions, such as opinion distribution, network homophily, modularity and LLM injection points.
Additionally, we observe that out-of-the-box LLMs, i.e., without fine-tuning via our methodology, fail to tailor the generation to the underlying network and opinion structures. This limitation is especially pronounced when the desired sentiment skews toward negative emotions, highlighting the necessity and effectiveness of our fine-tuning approach.

As a point of further discussion, it is important to note that our procedure is parametric to the engagement model, allowing customization to address the specific requirements of the user. The simulated feedback is represented as a scalar value corresponding to the number of activated users, making it compatible with outputs from other opinion dynamics or information diffusion models.
This flexibility also opens new avenues for future work, such as exploring how different assumptions on the engagement model influence the content generated by the LLM. For instance, analyzing the impact of zealots~\cite{brooks2024emergence}, stubborn users~\cite{ghaderi2014opinion}, and backfire mechanisms~\cite{jager2005uniformity} within the engagement model could reveal patterns in how the LLM balances the sentiment of the generated content.

\noindent{\textbf{Limitations.}} First, our experiments were limited to prompt completion. Nevertheless,  alternative approaches (e.g., generation from scratch) are possible. Further, the effects of crafting the prompt structure are not considered here and are worth being investigated. As also highlighted in the current literature for different tasks~\cite{salinas2024butterflyeffectalteringprompts}, approaches based on RAG~\cite{gao2024retrievalaugmentedgenerationlargelanguage} could also be adopted for proper crafting of the prompt and helping accurate response generation. 

\noindent{\textbf{Societal impact.}} We acknowledge the potential misuse of our findings, such as leveraging engagement optimization for harmful purposes like automating information operations~\cite{ferrara2016rise, weedon2017information}
using Generative AI~\cite{yang2024anatomy, cinus2024exposing}.
This is a common concern across literature focused on optimizing social objectives~\cite{kempe2003maximizing, barbieri2014influence, tu2022viral}. However, we argue that our approach holds significant promise for fostering social good. Recent studies demonstrate how LLMs can be utilized to find common ground in democratic deliberation~\cite{tessler2024ai}, incentivize the use of reliable news sources~\cite{askari2024incentivizing}, and de-radicalize extremist groups on platforms like Telegram~\cite{russo2024ai}. 
In this context, alternative modeling choices, such as optimizing reward functions to minimize network or opinion polarization, could further enhance the societal benefits of our framework.



\section*{Acknowledgements}
This work has been supported by the project SERICS (PE00000014) under the MUR National Recovery and Resilience Plan funded by the European Union -- NextGenerationEU, and by MUR on D.M.\ 351/2022, PNRR Ricerca, CUP H23C22000440007. Further, we acknowledge the CINECA award (HP10C9CXFF) under the ISCRA initiative for the availability of HPC resources.

\clearpage
\bibliographystyle{ACM-Reference-Format}
\bibliography{references}

\appendix

\section{Engagement-Sentiment Correlation}

\begin{figure}[ht!]
    \centering
    \includegraphics[width=0.32\columnwidth]{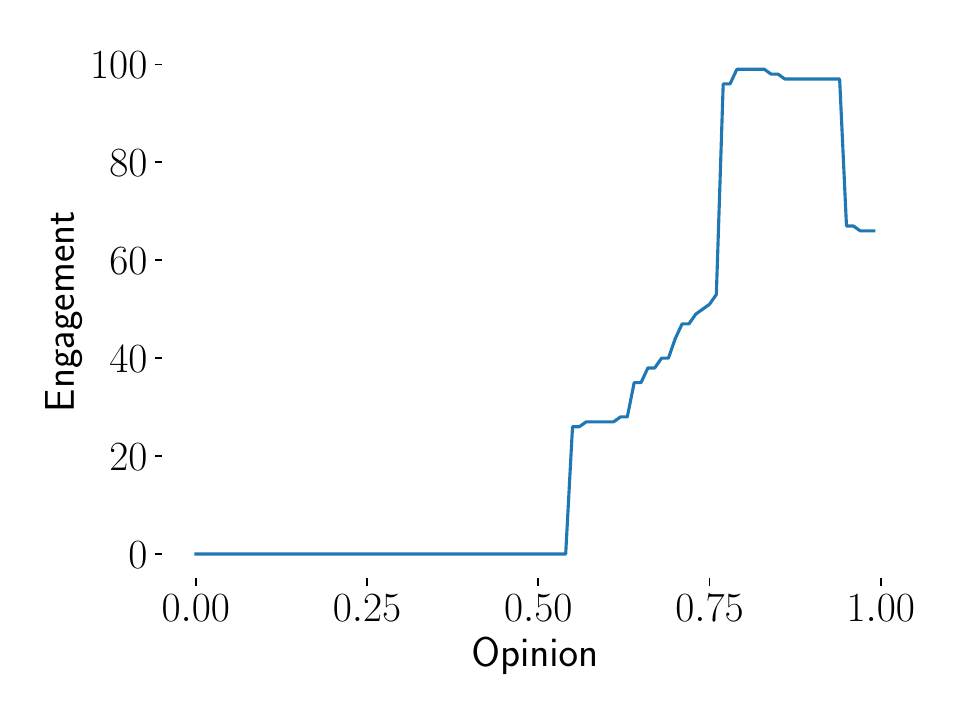}
    \includegraphics[width=0.32\columnwidth]{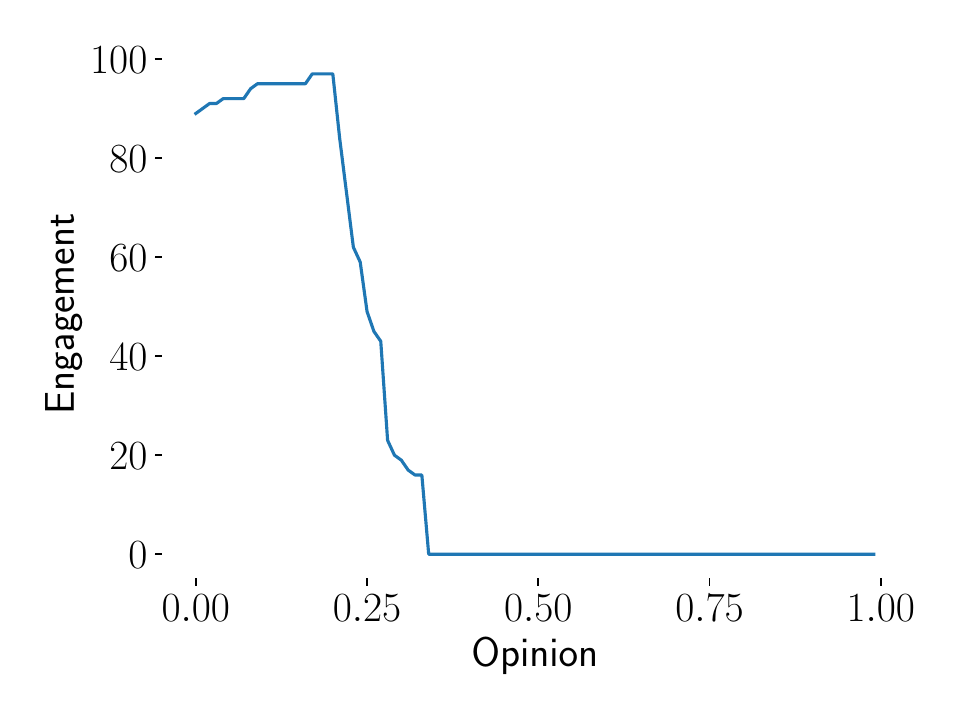}
    \includegraphics[width=0.32\columnwidth]{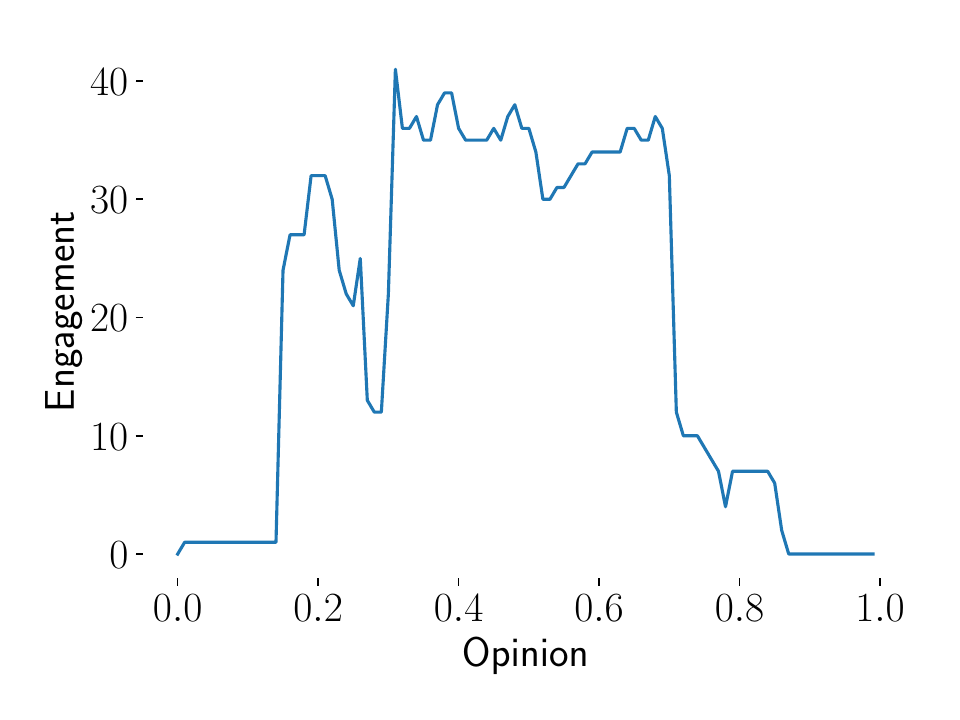} 
    \includegraphics[width=0.32\columnwidth]{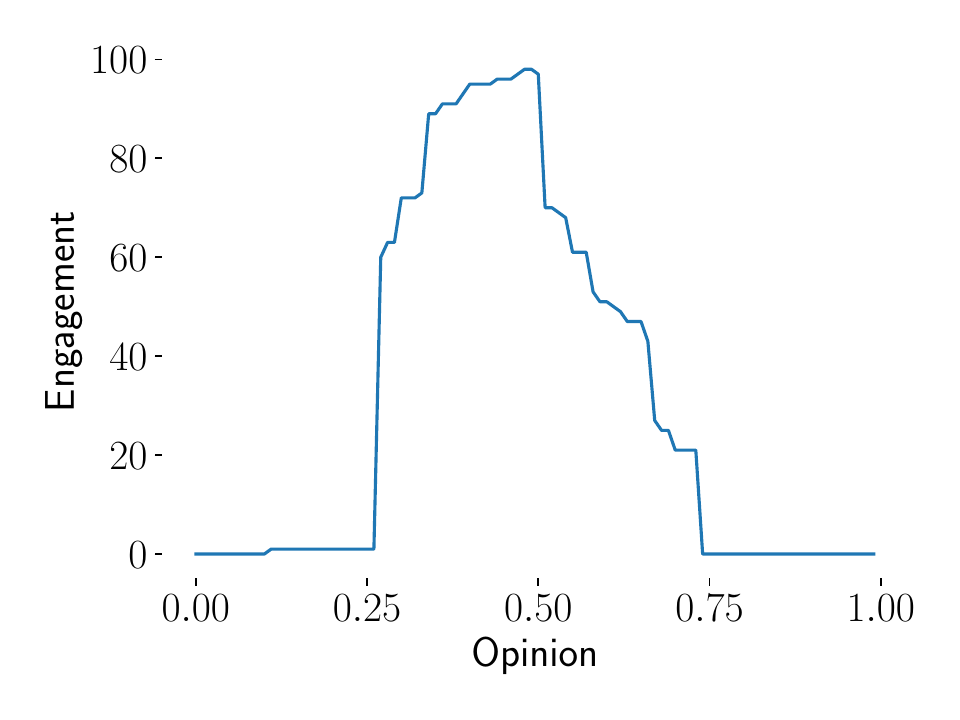}
    \includegraphics[width=0.32\columnwidth]{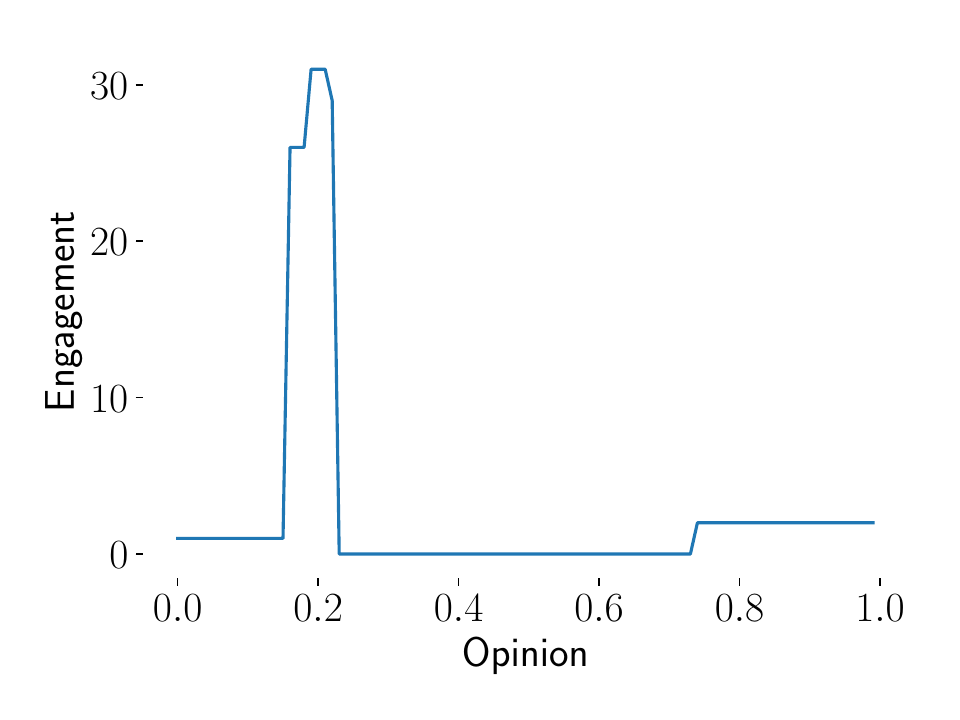}
    \caption{Engagement resulting from $\mathcal{M}_{\epsilon}$ varying the message sentiment $s_\content$. From top-left to bottom-right: network's opinion positive/negative/uniform/neutral/uniform; modularity high/low/high/low/high; homophily: high/low/low/high/low; position: central/comm-smallest/comm-largest/echo-high/echo-low.}
    \label{fig:sent-network-response}
\end{figure}

Figure~\ref{fig:sent-network-response} illustrates specific examples of engagement resulting from the propagation protocol $\propagationprotocol_\epsilon$ over the synthetic network, as the content sentiment $s_\content$ varies.
The plots are arranged from top-left to bottom-right. 
The first plot corresponds to a network with positively distributed opinions, high modularity, and strong homophily, where the source node is centrally positioned. In this case, engagement grows non-monotonically, reaching higher values for strong positive opinions (i.e., sentiment). The second plot represents a negatively distributed network with low modularity and weak homophily, with the source node placed in the smallest community (\texttt{Comm-smallest}). Here, engagement decreases as opinions become more positive, following an inverse trend compared to the first case.
The third plot depicts engagement in a network with a uniformly distributed opinion landscape, high modularity, and low homophily, where the source node is located in the largest community (\texttt{Comm-largest}). In this scenario, engagement remains at half of the maximum across a wide opinion range (0.15 to 0.85) and drops to zero elsewhere. The fourth plot corresponds to a neutral network with low modularity and high homophily, where the source node is positioned within a highly opinionated echo chamber (\texttt{Echo-high}), meaning it is inside the region with the highest average opinion. Here, the engagement range shrinks to 0.25–0.75, while its maximum value increases up to 100. Finally, the last plot represents engagement over a uniformly distributed network with high modularity and low homophily, where the source node is in the low-opinion echo chamber (\texttt{Echo-low}). In this case, engagement peaks around the dominant opinion within the echo chamber.

\section{Simulated Against Real-World Engagement}

\begin{figure}[ht!]
    \centering
    \includegraphics[width=0.8\columnwidth]{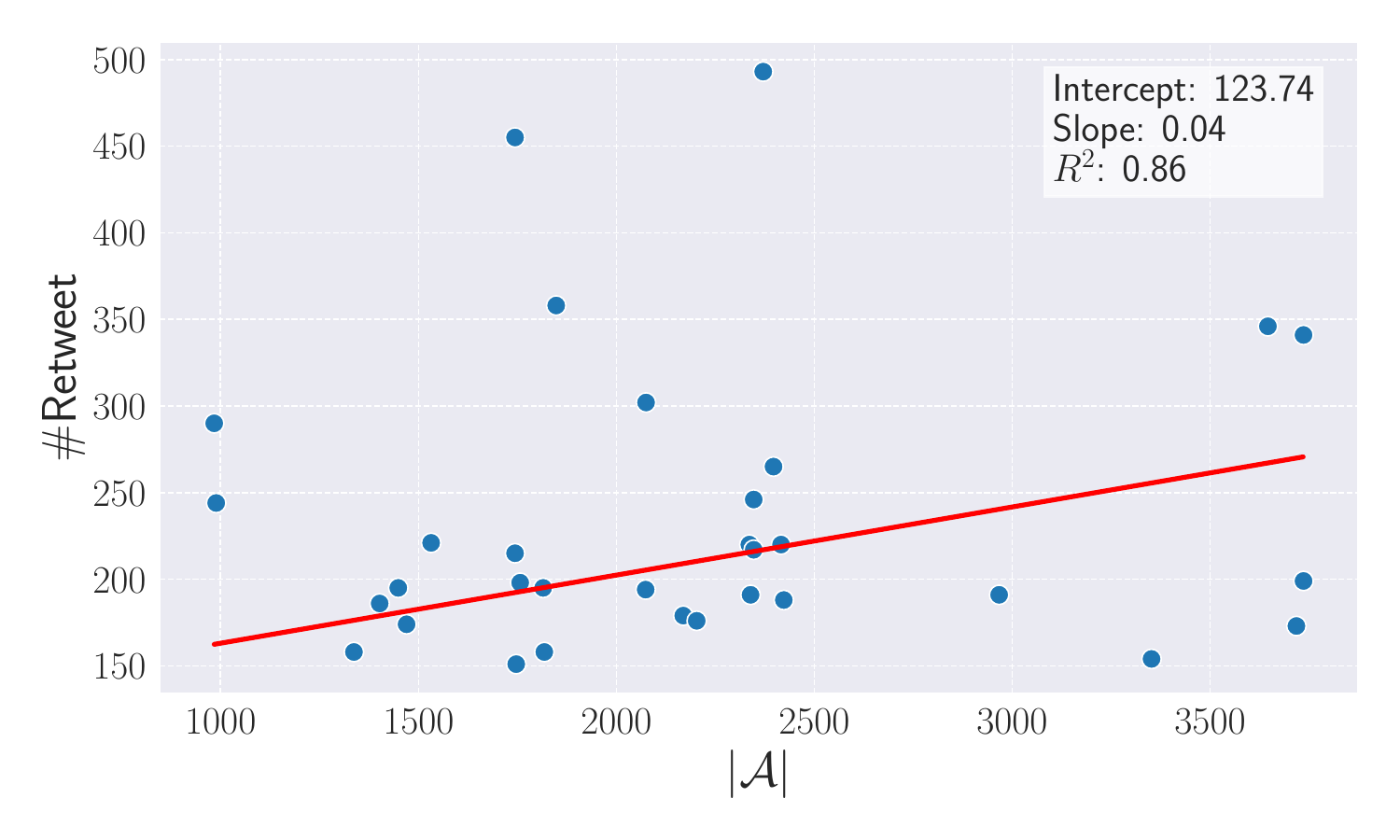}
    \caption{RANSAC regression line fitted on the number of active users $\lvert \mathcal{A} \rvert$ estimated by our engagement model versus the number of retweets of the most \emph{engaging} tweets in the Brexit dataset.}
    \label{fig:correlation-engagement-brexit}
\end{figure}
We validate the reliability of our propagation model by comparing the estimated engagement over the \textit{Brexit} real network with the actual number of retweets generated by the most engaging content in the dataset. Figure~\ref{fig:correlation-engagement-brexit} presents the corresponding regression line, computed using the Random Sample Consensus (RANSAC) method~\cite{CantzlerRandomSC}. Although our propagation protocol tends to overestimate engagement, we observe a positive correlation between the number of active users predicted by our model ($|\mathcal{A}|$) and the actual retweet count. This correlation supports the validity of our findings on the real network.

\end{document}